\documentclass{article}
\usepackage[final,dandb]{neurips_2025}
\usepackage{graphicx} \graphicspath{{figures/}}
\usepackage{amsmath,amssymb,mathabx,mathtools,amsthm,nicefrac}
\usepackage[pagebackref,breaklinks,colorlinks]{hyperref}
\usepackage[capitalise,noabbrev,nameinlink]{cleveref}
\usepackage[skip=3pt,font=small]{subcaption}
\usepackage[skip=3pt,font=small]{caption}
\usepackage{acronym}
\usepackage{wrapfig}
\usepackage[dvipsnames,svgnames,x11names,table]{xcolor}
\usepackage{enumerate,enumitem}
\usepackage{xspace}
\usepackage[misc]{ifsym}
\usepackage{verbatim,fancyvrb,listings}
\usepackage{booktabs,tabularx,colortbl,multirow,multicol,array,makecell,tabularray}
\usepackage{siunitx}
\usepackage{dblfloatfix}

\makeatletter
\DeclareRobustCommand\onedot{\futurelet\@let@token\@onedot}
\def\@onedot{\ifx\@let@token.\else.\null\fi\xspace}
\def\eg{\emph{e.g}\onedot} 

\def\ie{\emph{i.e}\onedot}

\makeatother

\newcommand{\dataset}[0]{\texttt{DrivAerStar}\xspace}
\newcommand{\olddataset}[0]{\texttt{DrivAer}\xspace}
\newcommand{\starccm}[0]{STAR-CCM+\textsuperscript{\textregistered}\xspace}
\newcommand{\blender}[0]{Blender\textsuperscript{\textregistered}\xspace}
\newcommand{\openfoam}[0]{OpenFOAM\textsuperscript{\textregistered}\xspace}
\acrodef{ai}[AI]{Artificial Intelligence}
\acrodef{cfd}[CFD]{Computational Fluid Dynamics}
\acrodef{pde}[PDE]{Partial Differential Equation}
\acrodef{cad}[CAD]{Computer Aided Design}
\acrodef{ffd}[FFD]{Free Formed Deformation}
\acrodef{bev}[BEV]{Battery Electric Vehicle}
\acrodef{hev}[HEV]{Hybrid Electric Vehicle}
\acrodef{icv}[ICV]{Internal Combustion Vehicle}
\acrodef{dns}[DNS]{Direct Numerical Simulation}
\acrodef{les}[LES]{Large Eddy Simulation}
\acrodef{ffd}[FFD]{Free Form Deformation}
\acrodef{lhs}[LHS]{Latin Hypercube Sampling}
\acrodef{rans}[RANS]{Reynolds-Averaged Navier–Stokes}
\acrodef{amg}[AMG]{Algebraic Multigrid}
\acrodef{piv}[PIV]{Particle Image Velocimetry}
\acrodef{sst}[SST]{Shear-Stress Transport}
\acrodef{simple}[SIMPLE]{Semi-Implicit Method for Pressure-Linked Equations}
\acrodef{license}[CC BY-NC-SA 4.0]{Creative Commons Attribution-NonCommercial-ShareAlike 4.0 International}

\newcommand{\company}[1]{\textsc{#1}\@} 
\newcommand{\round}[1]{ \num[round-mode=places,round-precision=4]{#1}} 

\newcommand{\homepagelink}[0]{\href{https://drivaerstar.github.io/}}

\title{\dataset: An Industrial-Grade \acs{cfd} Dataset for Vehicle Aerodynamic Optimization}

\author{%
    \normalfont\setlength{\tabcolsep}{3pt}%
    \hspace{-1em}\begin{tabular}{ccc}
        \textbf{Jiyan Qiu}$^{\,1}$ & \textbf{Lyulin Kuang}$^{\,1}$ & \textbf{Guan Wang}$^{\,2}$ \\
        \texttt{\footnotesize{}jiyanq0430@gmail.com} & \texttt{\footnotesize{}lkuang@nvidia.com} & \texttt{\footnotesize{}wangguan.nantes@gmail.com}
    \end{tabular}
    \vspace{6pt}\\
    \hspace{-1em}\begin{tabular}{ccc}
        \textbf{Yichen Xu}$^{\,3,1,4,5}$ & \textbf{Leiyao Cui}$^{\,3,4,5}$ & \textbf{Shaotong Fu}$^{\,1}$\\
        \texttt{\footnotesize{}xuyc@stu.pku.edu.cn} & \texttt{\footnotesize{}cuileiyaony@gmail.com} & \texttt{\footnotesize{}shaotongf@nvidia.com}
    \end{tabular}
    \vspace{6pt}\\
    \hspace{-1em}\begin{tabular}{cc}
        \textbf{Yixin Zhu}$^{\,3,4,5,6\,\textrm{\Letter}}$ & \textbf{Ruihua Zhang}$^{\,1,\,\textrm{\Letter}}$ \\
        \texttt{\footnotesize{}yixin.zhu@pku.edu.cn} & \texttt{\footnotesize{}ritaz@nvidia.com}
    \end{tabular}
    \vspace{6pt}\\
    \footnotesize$^{\textrm{\Letter}}$ Corresponding authors\quad{}
    \footnotesize$^1$ NVIDIA\quad{}
    \footnotesize$^2$ Baidu, \company{Inc.}\quad{}
    \footnotesize$^3$ Peking University\\
    \footnotesize$^4$ State Key Lab of General AI, Peking University\\
    \footnotesize$^5$ Beijing Key Laboratory of Behavior and Mental Health, Peking University\\
    \footnotesize$^6$ Embodied Intelligence Lab, PKU-Wuhan Institute for Artificial Intelligence
    \vspace{6pt}\\
    \url{https://drivaerstar.github.io/}
    \vspace{-12pt}
}

\begin{document}
\maketitle

\begin{abstract}
Vehicle aerodynamics optimization has become critical for automotive electrification, where drag reduction directly determines electric vehicle range and energy efficiency.
Traditional approaches face an intractable trade-off: computationally expensive \ac{cfd} simulations requiring weeks per design iteration, or simplified models that sacrifice production-grade accuracy. 
While machine learning offers transformative potential, existing datasets exhibit fundamental limitations---inadequate mesh resolution, missing vehicle components, and validation errors exceeding 5\%---preventing deployment in industrial workflows.
We present \dataset, comprising 12,000 industrial-grade automotive \ac{cfd} simulations generated using \starccm software. The dataset systematically explores three vehicle configurations through 20 \ac{cad} parameters via \ac{ffd} algorithms, including complete engine compartments and cooling systems with realistic internal airflow.
\dataset achieves wind tunnel validation accuracy below 1.04\%---a five-fold improvement over existing datasets---through refined mesh strategies with strict wall $y^+$ control. Benchmarks demonstrate that models trained on this data achieve production-ready accuracy while reducing computational costs from weeks to minutes.
This represents the first dataset bridging academic machine learning research and industrial \ac{cfd} practice, establishing a new standard for data-driven aerodynamic optimization in automotive development.
Beyond automotive applications, \dataset demonstrates a paradigm for integrating high-fidelity physics simulations with \ac{ai} across engineering disciplines where computational constraints currently limit innovation.
\end{abstract}

\section{Introduction}

Aerodynamic optimization has become increasingly crucial as the automotive industry undergoes rapid transformation toward environmentalization, electrification, and intelligentization \citep{bibra2022global}. Precise aerodynamic design directly impacts electric vehicle range \citep{huluka2019numerical}, fuel efficiency \citep{abinesh2014cfd}, fluid-excited noise \citep{wang2013hybrid}, and high-speed stability \citep{brandt2022high}. Beyond industrial applications, high-fidelity external flow field data enables deeper integration between \ac{cfd} and data-driven methods for studying complex flow dynamics. However, traditional numerical simulations require prohibitive computational resources, while existing datasets lack engineering fidelity for \ac{icv} \citep{payri2015challenging}, \ac{hev} \citep{hannan2014hybrid}, and \ac{bev} \citep{upadhyaya2023overview,bjerkan2016incentives} applications, creating urgent demand for datasets combining precision with comprehensive parameter coverage.

Recent automotive datasets like DrivAerNet++ \citep{elrefaie2024drivaernetpp} and DrivAerML \citep{ashton2024drivaerml} provide valuable foundations but exhibit critical limitations preventing industrial adoption. Their mesh strategies inadequately resolve gradient-dependent fields, causing boundary layer distortion and numerical diffusion errors. These resources focus on limited vehicle model deformations, failing to represent aerodynamic characteristics across diverse vehicle types and operating conditions. Furthermore, they deviate from industrial standards in turbulence modeling and solver configuration, producing results that diverge significantly from wind tunnel measurements. These shortcomings prevent the development of industrial-grade surrogate models capable of optimizing design features within minutes rather than weeks.

\begin{figure}[t!]
    \centering
    \includegraphics[width=\linewidth]{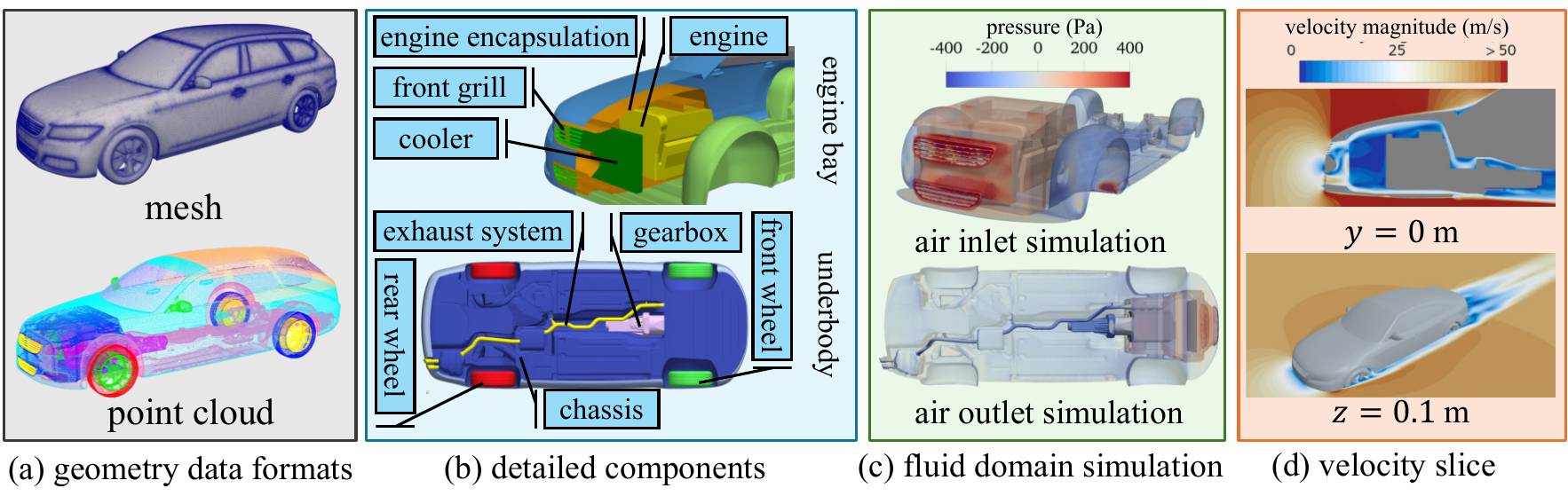}
    \caption{\textbf{\dataset dataset features.} (a) Multiple geometry data formats: mesh (top) serves as primary representation while point cloud (bottom) enables alternative computational approaches and geometry analysis. (b) Complete vehicle modeling includes detailed engine bay compartment assembly (top) and underbody components (bottom), distinguishing \dataset from previous automotive datasets. (c) Internal airflow simulation capabilities featuring air inlet (top) and outlet (bottom) simulations, enabled by precise internal component modeling to support comprehensive automotive system research. (d) Velocity field cross-sections showing internal engine compartment flow at vehicle centerline ($y=0$ m) and external flow field at tire axle plane ($z=0.1$ m).}
    \label{fig:geometry}
\end{figure}

We present \dataset, addressing these limitations through industrial-grade \starccm simulations. Our approach overcomes \openfoam's inherent wall shear stress calculation deficiencies by implementing refined mesh strategies with targeted aerodynamic sensitivity refinements, sub-regional surface adaptation, and optimized boundary layer resolution. Strict control of dimensionless wall distance $y^+$ values \citep{schlichting1961boundary} ensures accurate boundary layer simulation while achieving superior computational precision with comparable mesh counts. The dataset spans multiple vehicle platforms with systematic \ac{ffd}-based geometric parameter variations, representing diverse aerodynamic characteristics across vehicle types and operating conditions. Critically, we include complete front compartment assemblies with engine components and cooling systems, featuring realistic airflow channels that capture real-world vehicle aerodynamics.

Our 1,080,000 CPU core-hour computational investment across 100 nodes (see \cref{supp:sec:compute_cost}) produces a 20 TB dataset containing 12,000 samples---50\% more than existing automotive datasets. The comprehensive data covers pressure distributions, velocity vectors, turbulent kinetic energy, and high-precision aerodynamic coefficients with drag coefficient accuracy of $\pm 0.012$. Following Loughborough University Wind Tunnel Lab specifications \citep{varney2020experimental} and automobile manufacturer simulation standards, we achieve pressure coefficient errors below 1.04\% compared to experimental data. \cref{tab:dataset_comparison} presents detailed comparisons with existing automotive \ac{cfd} datasets.

\begin{table}[b!]
    \centering  
    \small
    \caption{\textbf{Comparative analysis of state-of-the-art automotive \acs{cfd} datasets.} Quantitative benchmarking across experimental validation capabilities, numerical resolution parameters, and simulation accuracy metrics. \acs{piv} indicates flow field validation availability; wall $y^+$ represents dimensionless wall distance for boundary layer resolution quality; average $C_D$ precision measures uncertainty bounds of drag coefficients; wind tunnel error shows percentage deviation from experimental measurements. Data are sourced from original literature and publicly released datasets, with detailed information available in \cref{supp:sec:vsbaseline}.}
    \label{tab:dataset_comparison}
    \resizebox{\linewidth}{!}{%
        \begin{tabular}{lccc}
            \toprule  
            \textbf{Metric} & DrivAerNet++ \scriptsize \citep{elrefaie2024drivaernetpp} & DrivAerML \scriptsize \citep{ashton2024drivaerml} & \dataset \\
            \midrule  
            PIV        & No    & No & \textbf{Yes} \\
            Engine compartment         & No    & No & \textbf{Yes} \\ 
            Cooler    & No    & No & \textbf{Yes} \\
            Samples              & 8,000                & 500             & \textbf{12,000} \\
            Wall $y^+$ range & $[50,300]$               & -            & $\mathbf{[30,200]}$ \\  
            Average $C_D$ precision     & 0.012          & 0.010       & \textbf{0.005} \\  
            Wind tunnel error (\%)    & \textbf{$\leq$5.0}                 & $\leq$6.5             & \textbf{1.04} \\  
            \bottomrule  
        \end{tabular}%
    }%
\end{table}

To validate effectiveness, we benchmark multiple machine learning architectures, including Transolver \citep{wu2024transolver}, GNOT \citep{hao2023gnot}, and PointNet \citep{qi2017pointnet}. Our evaluation analyzes surface pressure distributions and derived aerodynamic coefficients through surface pressure integration, establishing a multi-scale assessment for complete aerodynamic performance characterization. These benchmarks demonstrate \dataset's utility for training industrial-grade surrogate models while providing foundations for optimization research.

Our contributions are:
\begin{itemize}[leftmargin=*,noitemsep,nolistsep]
    \item \textbf{High-fidelity dataset}: \dataset provides 20 TB of external flow field data from 12,000 industrial-grade \starccm simulations across multiple vehicle configurations, delivering unprecedented data quality for surrogate model training.
    \item \textbf{Adaptive mesh methodology}: We employed refined mesh strategies with adaptive regional refinement and strict wall $y^+$ control, achieving superior boundary layer resolution that overcomes limitations of previous approaches.
    \item \textbf{Complete vehicle modeling}: Integration of front compartment assemblies with engines and cooling components creates continuous cooling flow channels, accurately representing real-world vehicle operating conditions.
    \item \textbf{Comprehensive benchmark}: We established multi-scale performance evaluation across surface and volumetric data, revealing current limitations and identifying research directions including scientific function discovery \citep{tang2023discovering}, forward \ac{pde} problems \citep{lu2021learning,raissi2019physics}, simulation super-resolution \citep{kochkov2021machine}, and inverse \ac{pde} reconstruction \citep{gao2021particle}.
\end{itemize}

\section{Related Work}

\paragraph{Machine Learning \acs{cfd} Benchmarks}

Large-scale, high-fidelity datasets form the foundation for deep learning applications in \ac{cfd}. Recent standardized benchmarks support research in physical simulation and \ac{pde} modeling, including BubbleML \citep{hassan2023bubbleml} for multiphase flows, Lagrangebench \citep{toshev2023lagrangebench} for Lagrangian particle-based simulations, BLASTNet \citep{chung2022blastnet} and BLASTNet 2.0 \citep{chung2023turbulence} for turbulent flow datasets, and PDEBench \citep{takamoto2022pdebench} for comprehensive \acs{pde}-based physical systems. While these datasets provide valuable contributions, they predominantly feature simplified geometries and idealized conditions that fail to capture the complex geometries and multiphysics interactions encountered in industrial applications.

\paragraph{Aerospace Aerodynamics Datasets}

Aerospace applications have driven advances in parameterized aerodynamic optimization through datasets enabling efficient simulation of complex geometries. AirfRANS \citep{bonnet2022airfrans} provides a 2D airfoil database built using \openfoam with parametric meshing strategies for systematic NACA airfoil studies across operational ranges (Mach numbers 0.2--0.8, angles of attack from $-8^\degree$ to $20^\degree$), establishing clear mappings between geometric deformations and aerodynamic performance while providing insights into transonic shock effects. Extending to three dimensions, AircraftVerse \citep{cobb2023aircraftverse} offers 27,714 complete aircraft configurations with varied wing parameters, powertrain characteristics, and performance metrics, combining deep learning for geometry generation with engineering models for aerodynamic calculations. However, these aerospace datasets address fundamentally different flow regimes and geometric complexities than automotive applications, limiting their direct applicability to ground vehicle aerodynamics.

\begin{figure}[t!]
    \centering
    \includegraphics[width=\linewidth]{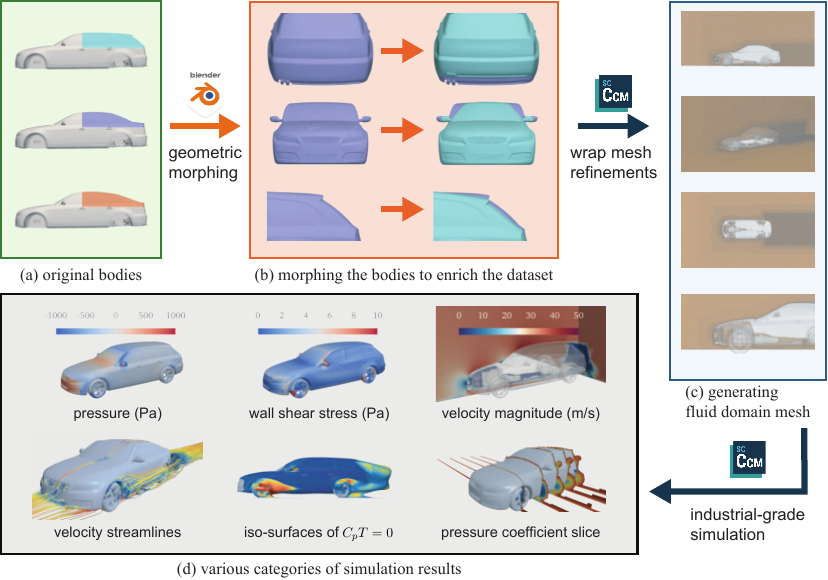}
    \caption{\textbf{\dataset data generation.} (a) Three canonical \olddataset reference bodies (Estateback, Notchback, and Fastback) serve as geometric foundations. (b) Parametric morphing systematically varies 20 vehicle components, including greenhouse (top), rear diffuser (middle), and trunk lid (bottom). (c) Industrial-grade mesh generation in \starccm produces refined hexahedral-dominant meshes with precise wheel alignment and boundary layer resolution for complex flow capture. (d) Comprehensive \acs{cfd} simulations generate diverse flow visualizations: surface pressure, wall shear stress, velocity magnitude, streamline patterns, flow separation regions ($C_P=0$ iso-surfaces), and pressure coefficient slices revealing key aerodynamic structures.}
    \label{fig:pipeline}
\end{figure}

\paragraph{Automotive Aerodynamics Datasets}

Early automotive datasets based on simplified geometries---ShapeNet Car \citep{umetani2018learning,song2023shapenetcar}, WindSor Car \citep{ashton2024windsorml}, and Ahmed-body configurations \citep{li2023gino,ashton2024ahmedml}---advanced the field but created substantial gaps between research outcomes and industrial requirements. More sophisticated approaches emerged with the DrivAerNet series, including DrivAerNet \citep{elrefaie2025drivaernet} and DrivAerNet++ \citep{elrefaie2024drivaernetpp}, implementing parametric simulations using \openfoam. DrivAerML similarly offers 500 parametrically morphed variants of the \olddataset Notchback model with pressure, velocity, and turbulence fields.
Concurrent work by \citet{warner2025aerosuv} introduced AeroSUV, featuring 1,000 detached eddy simulation configurations from an open-access reference model extending the \olddataset platform, executed using proprietary in-house solvers. Despite these advancements, existing automotive datasets exhibit critical limitations preventing industrial adoption: insufficient mesh resolution for complex geometries, inadequate turbulence modeling within the \openfoam framework, and failure to account for crucial interactions between engine compartment thermal management and external aerodynamics.

\dataset addresses these fundamental limitations by employing \olddataset vehicle geometry with the industry gold standard \starccm solver for mesh generation and flow solution. As detailed in \cref{tab:dataset_comparison}, our approach delivers substantial advantages in mesh resolution, aerodynamic accuracy, parametric dimensionality, and engine compartment simulation fidelity. These advancements establish a new benchmark for data-driven automotive aerodynamics research, bridging the gap between academic machine learning exploration and practical engineering applications while supporting development of general pretrained aerodynamics operator networks with potential long-term impacts detailed in \cref{supp:sec:impact}.

\section{The \dataset Dataset}\label{sec:dataset_pipeline}

Developed using industry-standard \starccm, \dataset bridges academic research and industrial applications through rigorous validation, achieving 1.04\% mean relative error compared to wind tunnel measurements (\cref{table:cdvalidwind}). The dataset comprises 12,000 high-fidelity simulations of engine-integrated vehicle geometries, each containing STL representations with at least 4 million triangles and complementary data formats capturing surface distributions, full-field flow, and cross-sectional aerodynamic data. Our workflow (\cref{fig:pipeline}) encompasses geometric morphing (\cref{sec:geo_morph}), mesh generation (\cref{sec:mesh_gen}), and industrial-grade simulation (\cref{sec:sim}).

\begin{figure}[b!]
    \centering
    \includegraphics[width=\linewidth]{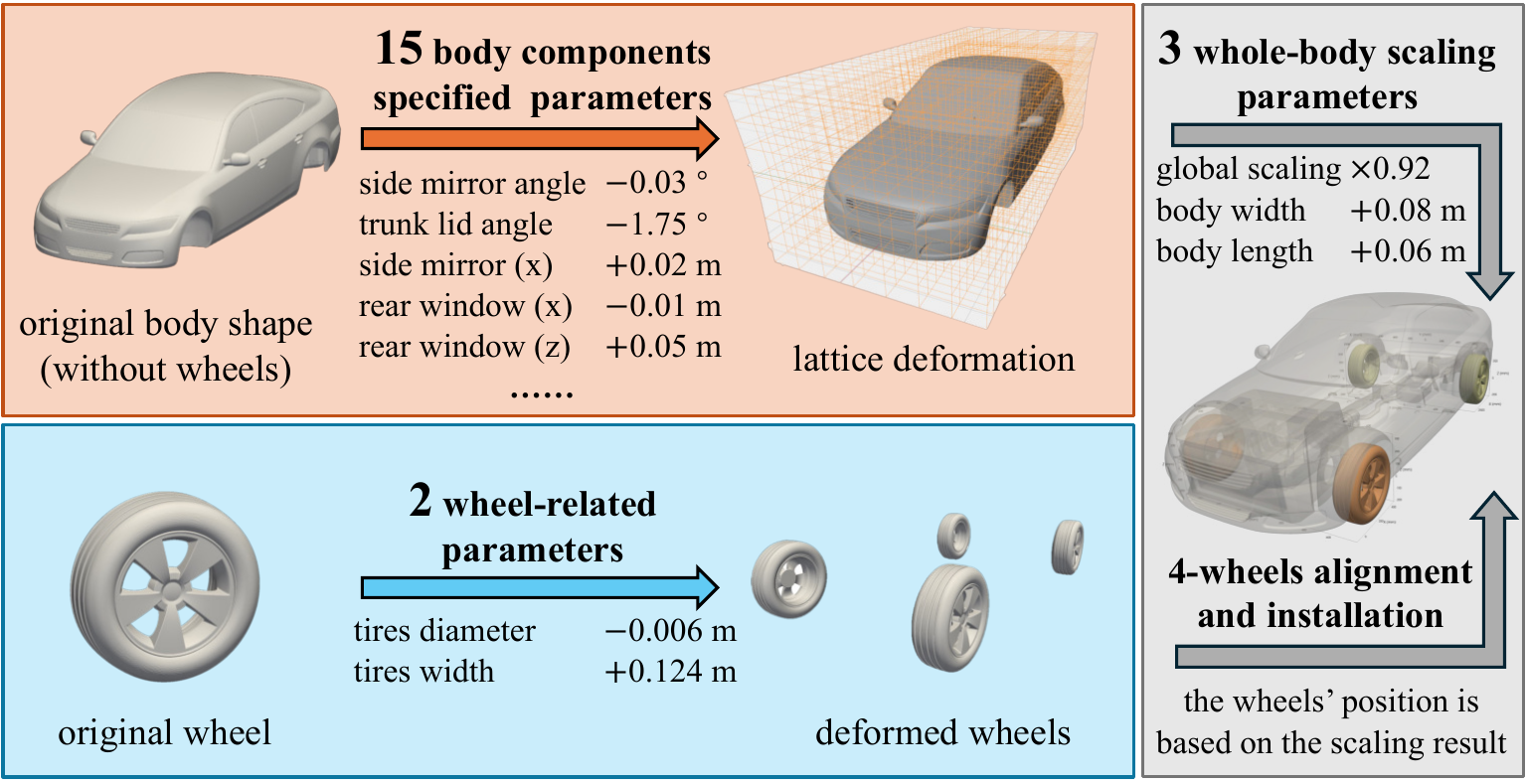}
    \caption{\textbf{Geometric morphing pipeline example.} Our parametric deformation framework transforms the baseline model through three sequential operations: (i) \textbf{Body component morphing} (top-left) applies 15 controlled parameter adjustments, including dimensional modifications and component repositioning; (ii) \textbf{Wheel morphing} (bottom-left) enables precise tire parameter control; and (iii) \textbf{Whole-body scaling and wheel installation} (right) applies 3 scaling parameters while aligning wheels to scaled positions. This systematic \acs{ffd} approach generates geometrically diverse yet aerodynamically realistic vehicle configurations.}
    \label{fig:morph_pipeline}
\end{figure}

Unlike existing automotive \ac{cfd} datasets, \dataset incorporates the highest-fidelity geometric data with three rear body configurations, each featuring complete air intake, engine compartment, and cooling systems (\cref{fig:geometry}). To ensure industrial relevance, we systematically varied 20 fine-tuned \ac{cad} parameters (\cref{tab:supp:car-parameters}) covering both local and global vehicle features using \blender through \ac{lhs}, creating a comprehensive design exploration space (\cref{supp:fig:deform}). The dataset spans Reynolds numbers from $9.46 \times 10^6$ to $1.48 \times 10^7$, with flow features available in volumetric, slice, and surface formats including velocity, pressure, surface properties, wall shear stress, mesh configurations, $y^+$ values, 3D streamlines, and iso-surfaces.

\dataset provides comprehensive aerodynamic metrics aligned with physics field data: pressure coefficient ($C_P$), friction coefficient ($C_F$), drag coefficient ($C_D$), and lift coefficient ($C_L$). Beyond performance metrics, we supply complete numerical data (pressure, velocity, wall shear stress) alongside reproducible workflows with validation scripts following industrial standards. Simulation results are detailed in \cref{supp:sec:sim_result}.
The dataset subset, encompassing over 12,000 VTK surface files and EnSight flow field files totaling approximately 20 TB, is accessible via our project \homepagelink{homepage} under the \ac{license} license, with data processing and benchmarking code included. Detailed licensing information is provided in \cref{supp:sec:licensing}.

\subsection{Geometric Morphing}\label{sec:geo_morph}

\dataset builds upon the experimentally validated \olddataset reference vehicle platform \citep{heft2012experimental,strangfeld2013experimental}, featuring three canonical rear configurations: Fastback, Estateback, and Notchback. To generate our comprehensive dataset of 12,000 geometrically diverse vehicle models, we implemented systematic \ac{ffd} techniques targeting critical aerodynamic components. This approach (\cref{fig:morph_pipeline}) enables precise parametric control over vehicle geometry, transforming the base Fastback model into varied body and wheel configurations through controlled parameter adjustments. Complete technical details of our \ac{ffd} implementation are provided in \cref{supp:sec:ffd}.

\subsection{Mesh Generation}\label{sec:mesh_gen}

Our automated mesh generation workflow consists of two sequential stages: surface wrapping and volume mesh generation.

\paragraph{Surface Wrapping}

Morphed geometries frequently exhibit topological defects, including mesh intersections and open edges, requiring systematic resolution. We employed \starccm's surface wrapping functionality on individual components---vehicle body, drivetrain, engine compartment, wheels, and coolers---to ensure simulation-ready geometry. Critical regions with fine features, particularly drivetrain and grille openings, received 10\% relative size refinement and specialized contact prevention parameters to maintain geometric fidelity while eliminating surface intersections. Each component underwent independent wrapping before integration into the computational domain. Complete technical specifications are detailed in \cref{supp:sec:surface_wrap}.

\paragraph{Volume Mesh Generation}

\begin{figure}[tb!]
    \centering
    \begin{subfigure}[t]{0.5\linewidth}
        \centering
        \includegraphics[width=\linewidth]{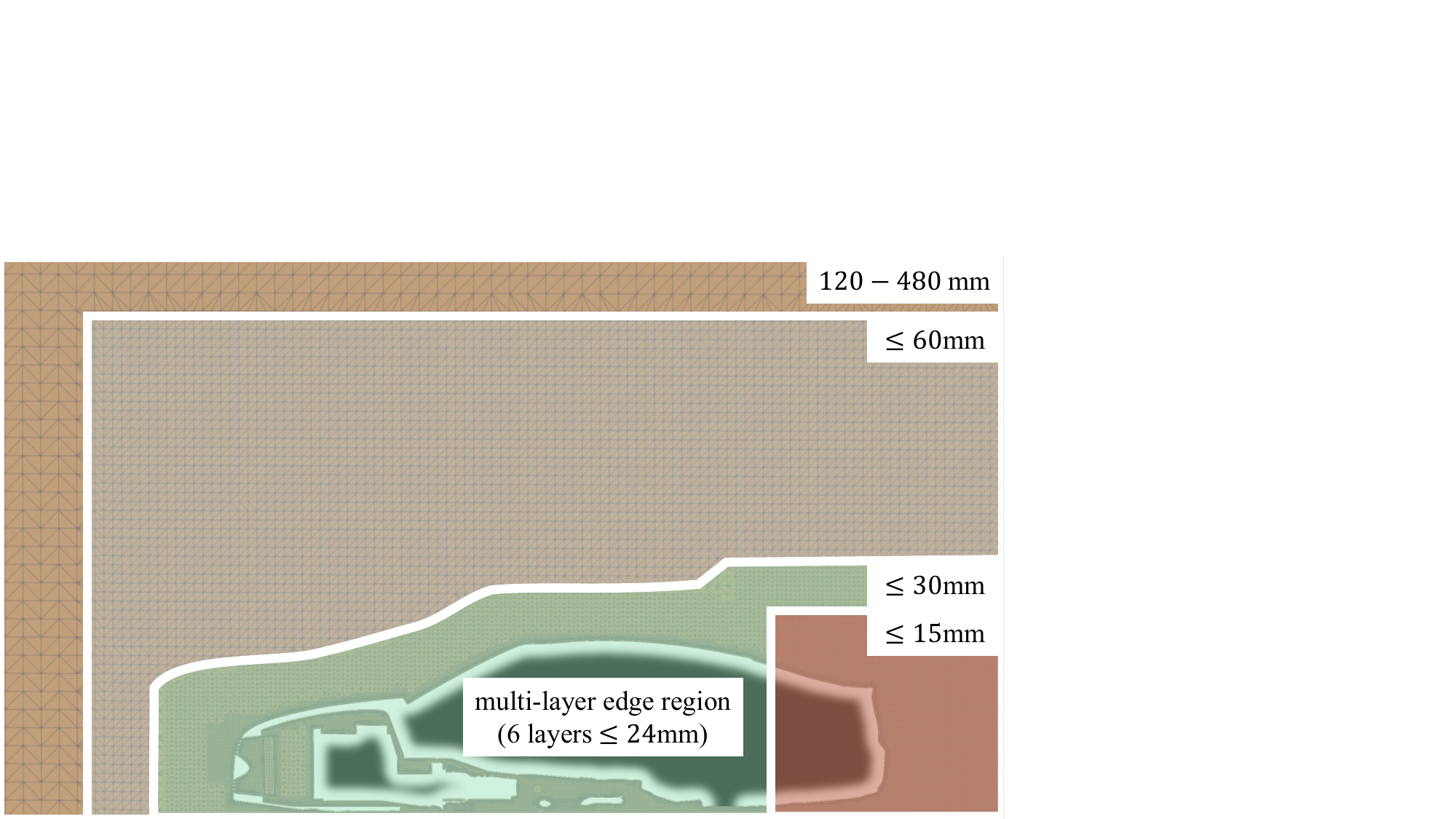}
        \caption{mesh slice at $y=0$ m}
    \end{subfigure}%
    \begin{subfigure}[t]{0.5\linewidth}
        \centering
        \includegraphics[width=\linewidth]{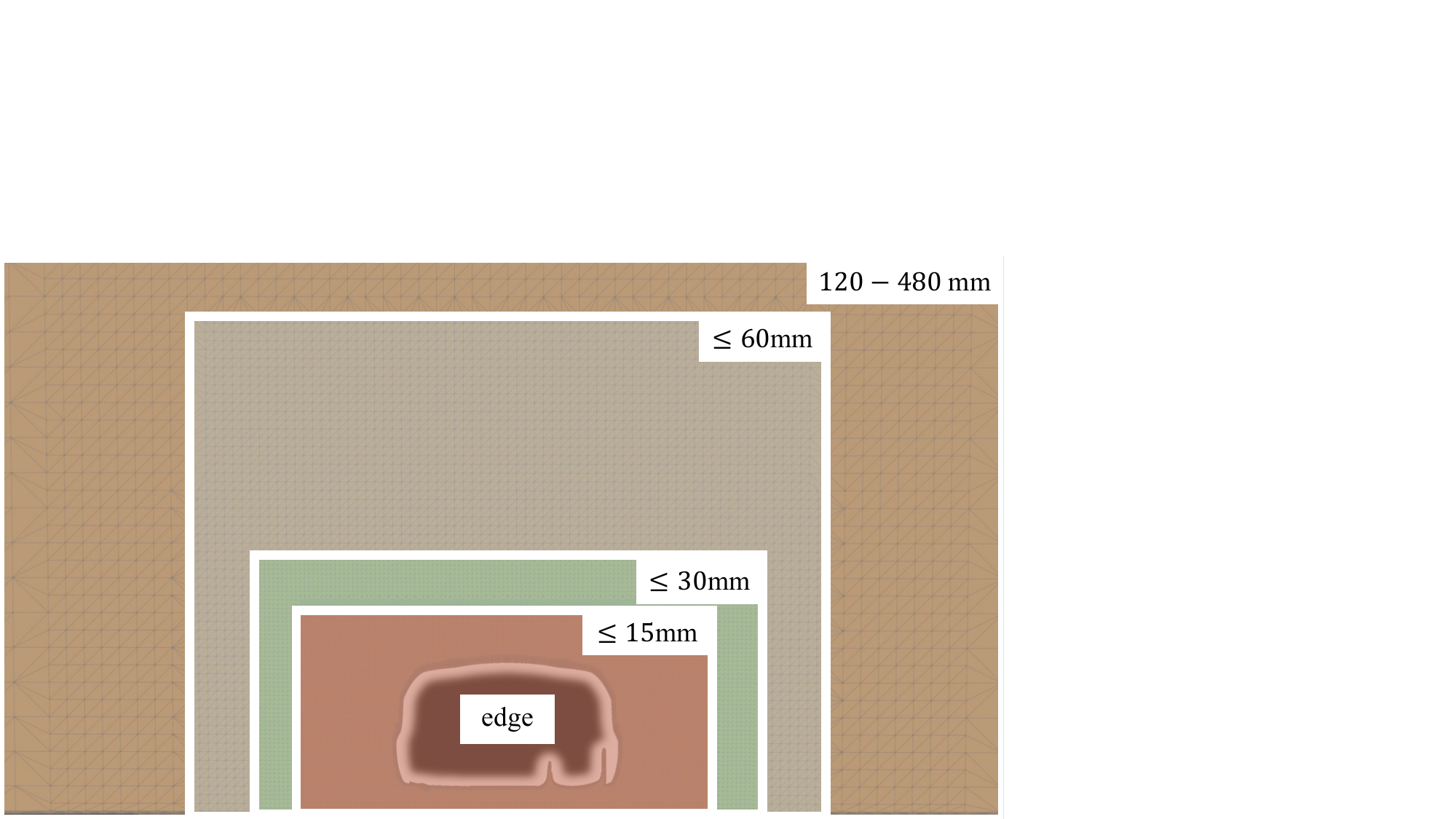}
        \caption{mesh slice at $x=3.12$ m}
    \end{subfigure}%
    \caption{\textbf{Cross-sectional views of mesh generation strategy.} (a) Longitudinal section at $y=0$ m shows four nested refinement zones with progressive cell density: far-field (120-480 mm), intermediate domain ($\leq$ 60 mm), near-body region ($\leq$ 30 mm), and high-resolution zone ($\leq$ 15 mm) surrounding the vehicle. Dedicated 6-layer boundary layer refinement with total thickness under 24 mm captures viscous phenomena. (b) Transverse section at $x=3.12$ m demonstrates consistent refinement strategy with graduated cell density approaching vehicle surfaces, enabling accurate wake structure resolution while maintaining computational efficiency.}
    \label{fig:mesh_gen}
\end{figure}

Our volume mesh (\cref{fig:mesh_gen}) employs refinement strategies aligned with validated practices \citep{heft2012experimental,zhang2019turbulence} using \starccm's production-grade algorithms. Each simulation domain comprises approximately 12 million hexahedral-dominant cells with strategic density distribution targeting aerodynamically significant features. The meshing protocol incorporates high-resolution prismatic boundary layers on all vehicle surfaces---both external and internal---ensuring proper resolution of boundary phenomena including separation, reattachment, and near-wall flow structures. To maintain dataset consistency, we automated the entire meshing sequence through Java macros with identical parameters across all 12,000 simulations. Additional details are provided in \cref{supp:sec:mesh}.

\subsection{Industrial-grade Simulation}\label{sec:sim}

\paragraph{Computational Domain and Boundary Conditions}

Our framework implements automotive industry-standard simulation protocols using a domain extending 15 vehicle lengths streamwise, 10 widths laterally, and 12 heights vertically. The inlet boundary, positioned three vehicle lengths upstream, imposes \SI{40}{m/s} (\SI{144}{km/h}) freestream velocity with 1\% turbulence intensity generated through spectral synthesizer algorithms. The downstream boundary enforces zero-gradient pressure outlet conditions. Vehicle-ground interaction employs sliding wall boundaries with rotating reference frames, calculating wheel angular velocity ($\omega = u_\infty/r$) based on instantaneous tire geometry. The engine compartment thermal management system incorporates experimentally validated porous media models with 0.6 porosity. All simulations utilize T/CSAE112-2019 standard atmospheric conditions: air density $\rho = \SI{1.225}{kg/m^3}$ and dynamic viscosity $\mu = \SI{1.85508e-5}{Pa\cdot s}$.

\paragraph{Flow Physics Modeling}

Simulations solve 3D steady-state \ac{rans} equations \citep{reynolds1895on} for incompressible flow with comprehensive wall distance modeling. We employed \ac{sst} \citep{menter1994two} $k$-$\omega$ turbulence formulation, selected for demonstrated accuracy in predicting flow separation and reattachment phenomena critical to automotive aerodynamics. Our hybrid all-$y^+$ wall treatment automatically transitions between direct viscous sublayer resolution and wall function approaches based on local mesh refinement, maintaining $y^+$ values below 200 while optimizing computational efficiency. Gradient reconstruction utilizes cell-based least squares minimization, ensuring numerical stability in regions with strong pressure gradients and separated flow. Complete governing equation formulations are provided in \cref{supp:sec:turbulence_models}.

\paragraph{Sovler}

Our solution strategy employs a pressure-based segregated solver with \ac{simple} algorithm \citep{patankar1983calculation} for pressure-velocity coupling. Spatial discretization implements second-order schemes for pressure and momentum equations, with first-order upwind treatment for turbulence quantities, enhancing stability. Linear equation systems are resolved using \ac{amg} methods \citep{xu2017algebraic} with V-cycle iterations (1 pre/post-sweep), achieving convergence to residuals below $10^{-5}$ within 30 iterations per time step. Carefully calibrated under-relaxation factors (0.7 for velocity, 0.8 for turbulence quantities) balance convergence rate with solution stability. Gauss-Seidel relaxation enhances iterative efficiency, particularly in high-gradient regions such as wheel wells and underbody flow channels.

\subsection{Experimental Validation}

\paragraph{Wind Tunnel Validation}

We validated our simulation methodology against high-fidelity measurements from Loughborough University Large Wind Tunnel facility \citep{johl2004design}, focusing on drag coefficient ($C_D$)---the critical automotive aerodynamics performance metric. Our 25\%-scale simulations precisely matched physical models provided by FKFS Stuttgart, ensuring geometric consistency between computational and experimental configurations. As shown in \cref{table:cdvalidwind}, simulations demonstrate exceptional agreement with experimental measurements across all three rear body configurations (Fastback, Notchback, Estateback), achieving 1.04\% average relative deviation with maximum discrepancy below 1.8\%. This correlation substantiates our simulation methodology's predictive accuracy for aerodynamic performance assessment. Detailed convergence analysis and mesh independence studies are presented in \cref{supp:sec:validation_quality}.

\begin{table}[ht!]
    \centering
    \small 
    \caption{\textbf{Data scaling effects on drag coefficient prediction accuracy.} Transolver model performance analysis with increasing training sample size. Left and center: Scatter plots of predicted versus ground truth $C_D$ values for models trained on 400 and 1200 samples, respectively, with diagonal green line representing perfect prediction ($y=x$) and red regression line showing actual prediction trends with correlation coefficients. Right: Violin plots comparing prediction error distributions, demonstrating improved performance with increased data volume (mean absolute error decreasing from 2.99\% to 2.78\%). Results confirm \dataset's effective learning scaling properties and potential for further accuracy improvements with additional training data. Comprehensive scaling results across all configurations are detailed in \cref{supp:sec:benchmark_results}.}
    \begin{tabular*}{\columnwidth}{@{\extracolsep{\fill}}lccc@{}}
        \toprule
        \textbf{Rear Configuration} & \textbf{Wind Tunnel $C_D$} & \textbf{\dataset $C_D$} & \textbf{Deviation (\%)} \\
        \midrule
        FastBack    & 0.278 & 0.2749  & 1.12\% \\
        NotchBack   & 0.279 & 0.2767  & 0.82\% \\
        Estateback  & 0.299 & 0.2955  & 1.17\% \\
        \bottomrule
    \end{tabular*}
    \label{table:cdvalidwind}
\end{table}

\paragraph{\acs{piv} Validation}

\begin{figure}[t!]
    \centering
    \includegraphics[width=\linewidth]{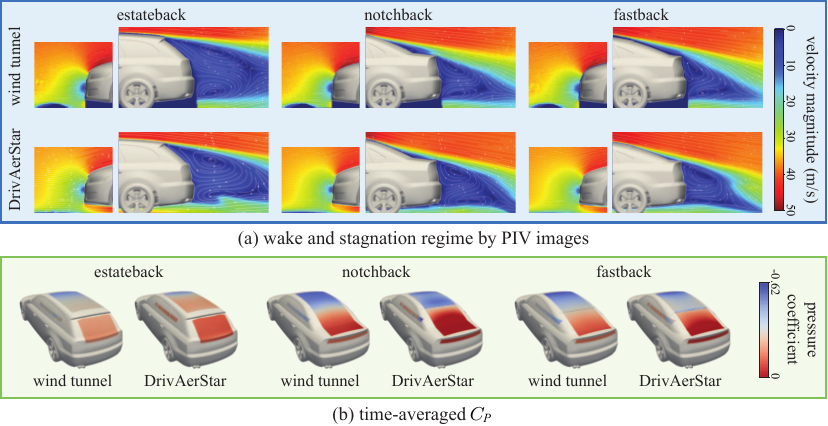}
    \caption{\textbf{Validation of flow physics predictions against wind tunnel measurements.} (a) Velocity magnitude distributions compare wind tunnel \acs{piv} measurements (top row) with \dataset predictions (bottom row) in wake regions and stagnation zones, demonstrating excellent flow structure agreement. (b) Surface pressure coefficient ($C_P$) distributions across vehicle body surfaces show consistent prediction accuracy between experimental and computational results for all three vehicle configurations, confirming accurate capture of complex aerodynamic phenomena.}
    \label{fig:piv}
\end{figure}

Flow structure prediction was further validated through \ac{piv} measurements in aerodynamically critical regions---front stagnation zones, A-pillar vortices, and rear wake structures. Surface pressure distributions were compared using data from 60 strategically positioned pressure taps, emphasizing the interchangeable rear configurations. As visualized in \cref{fig:piv}, these comparisons demonstrate excellent agreement in both magnitude and spatial distribution of pressure coefficients and velocity fields. The observed congruence confirms that our dataset accurately captures complex aerodynamic phenomena governing vehicle performance, including internal engine bay flows typically omitted in simplified models. Quantitative simulation-experiment difference analysis is provided in \cref{supp:sec:diffb}.

\section{Benchmarks on \dataset}

\subsection{Experimental Setup}

\paragraph{Benchmark Objectives}

We establish comprehensive machine learning benchmarks on the \dataset dataset to evaluate its industrial relevance and predictive capability. These benchmarks assess state-of-the-art deep learning architectures on aerodynamic prediction tasks using high-fidelity \ac{cfd} data representative of real automotive development. The primary task involves predicting surface pressure ($\tilde{p}$) and wall shear stress ($\tilde{\tau}_w$) distributions from vehicle geometry, as well as the derived drag coefficients ($C_D$). This setup aligns with recent advances in automotive aerodynamics machine learning research~\citep{hao2023gnot,wu2024transolver,nabian2024xmeshgraphnet,choy2025figconvnet,liu2025aerogto,bleeker2025neuralcfd} while providing realistic validation through our industry-standard dataset.

\paragraph{Dataset Configurations}

We systematically evaluated dataset scale and diversity effects. Initial benchmarks used restricted training sets (400-1200 samples) from single vehicle configurations (Estateback), while advanced experiments incorporated multi-configuration datasets spanning all three rear body types (Fastback, Estateback, Notchback). All experiments maintained consistent validation and test splits with 150 cases per configuration type. Additional experimental details are provided in \cref{supp:sec:exp_detail}.

\paragraph{Evaluation Protocol}

Model performance evaluation ensures accurate local flow physics representation and reliable global performance prediction---essential requirements for industrial applications. For spatial distribution accuracy of surface fields, we employ the relative $L_2$ error. Each flow field variable $\phi$ (\eg, velocity, pressure) is standardized via mean ($\mu_\phi$) and standard deviation ($\sigma_\phi$) computed from dataset samples (\ie, $\hat{\phi} = \frac{\phi - \mu_\phi}{\sigma_\phi}$):
$
    \epsilon = \frac{||\hat{\phi}_{\text{pred}} - \hat{\phi}_{\text{true}}||_2}{||\hat{\phi}_{\text{true}}||_2},
$
eliminating scale discrepancies among variables, where $\hat{\phi}_{\text{pred}}$ and $\hat{\phi}_{\text{true}}$ denote standardized predicted and ground-truth flow field variables, respectively. For integrated quantities (\eg, drag coefficient $C_D$), we assess absolute and percentage errors to ensure correct global performance prediction. Complete metric definitions and standardization procedures are provided in \cref{supp:sec:eval_metrics}.

\begin{figure}[t!]
    \centering
    \includegraphics[width=\linewidth]{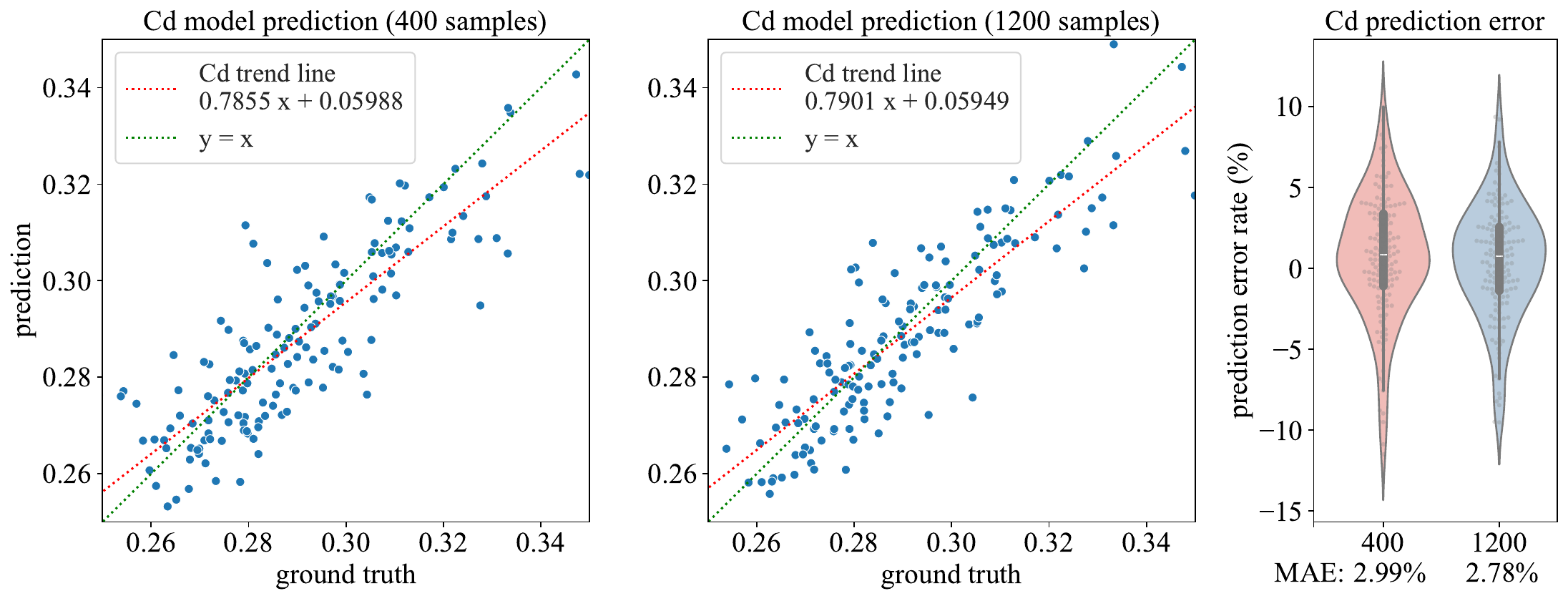}
    \caption{\textbf{Data scaling effects on drag coefficient prediction accuracy.} Analysis of Transolver model performance with increasing training sample size. Left and center: Scatter plots of predicted versus ground truth $C_D$ values for models trained on 400 and 1200 samples, respectively. The diagonal green line represents perfect prediction ($y=x$), while the red regression line shows the actual prediction trend with correlation coefficients. Right: Violin plots comparing prediction error distributions, showing improved performance with increased data volume (mean absolute error decreasing from 2.99\% to 2.78\%). This demonstrates the dataset's effective learning scaling properties and potential for further accuracy improvements with additional samples. Comprehensive scaling results across all configurations are provided in \cref{supp:sec:benchmark_results}.}
    \label{fig:results_gen}
\end{figure}

\subsection{Results} 

\paragraph{Surface Field Prediction}

We evaluated three state-of-the-art deep learning architectures on \dataset for predicting surface pressure and wall shear stress fields, as well as the derived drag coefficients. The performance metrics on the test set are summarized in \cref{tab:framework_comp}.

\begin{figure}[t!]
    \begin{subfigure}[t]{0.5\linewidth}
        \centering
        \includegraphics[width=\linewidth]{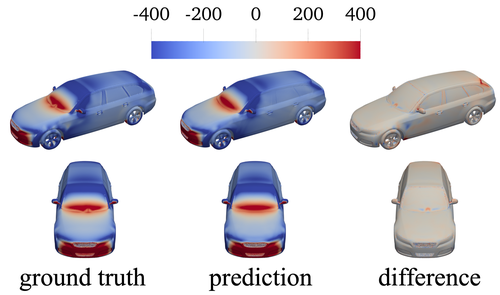}
        \caption{surface pressure (Pa)}
    \end{subfigure}%
    \hfill
    \begin{subfigure}[t]{0.5\linewidth}
        \centering
        \includegraphics[width=\linewidth]{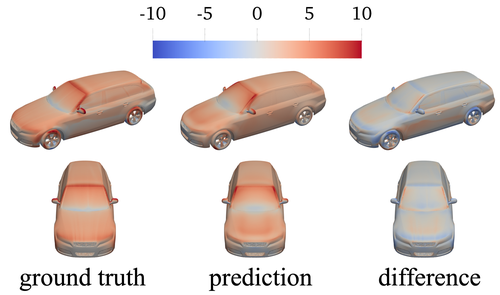}
        \caption{wall shear stress (x) (Pa)}
    \end{subfigure}%
    \caption{\textbf{Surface field prediction accuracy across critical aerodynamic regions.} (a) Surface pressure distributions (Pa) comparing ground truth \ac{cfd} simulation (left), neural network prediction (middle), and point-wise difference (right). (b) Wall shear stress $\tau_x$ (Pa) in streamwise direction with ground truth (left), model prediction (middle), and prediction error (right). The model accurately captures high-gradient regions around A-pillars and wheel wells while maintaining precision in uniform flow areas, demonstrating robust performance across diverse aerodynamic phenomena.}
    \label{fig:results_surface}
\end{figure}

Each model was trained on both specialized single-configuration subsets (Estateback only) and the complete multi-configuration dataset (Fastback, Notchback, and Estateback). Results consistently showed higher prediction errors on the complete dataset, confirming that morphological diversity significantly increases prediction complexity. This performance differential quantifies the fundamental trade-off between prediction accuracy and geometric variability in aerodynamic machine learning tasks. Representative prediction visualizations are presented in \cref{fig:results_surface}.

\paragraph{Data Scaling Analysis}

To investigate model generalization capabilities and data efficiency, we conducted systematic scaling experiments following \olddataset protocols. Both single-configuration subsets and complete datasets were divided into training (80\%) and test (20\%) partitions, with progressive increases in training samples while maintaining consistent test evaluation. The resulting learning curves (\cref{fig:results_gen}) reveal characteristic scaling behaviors for each architecture and quantify data requirements for achieving specific prediction accuracy thresholds.

We further assessed training size effects on Transolver's predictive performance for drag coefficient ($C_D$)---a key \ac{cfd} metric for vehicle aerodynamic optimization---in multi-vehicle settings. Results including $C_D$ validation loss and relative improvement values are summarized in \cref{table:cd_scaling}. Raw scaling experiment data (loss curves, test sample distributions) are provided in \cref{supp:sec:scaling}.

\begin{table}[t!]
    \centering
    \small  
    \setlength{\tabcolsep}{25pt}
    \caption{\textbf{Drag coefficient prediction scaling analysis in multi-vehicle configurations.} Transolver model performance across varying training dataset sizes (400–12,000 samples) evaluated on a consistent validation set of 150 samples per vehicle configuration. Learning curves demonstrate systematic improvement in $C_D$ prediction accuracy with increased training data, confirming the dataset's scalability and the model's ability to leverage additional samples for enhanced aerodynamic performance prediction across diverse vehicle geometries.}
    \label{table:cd_scaling}
    \begin{tabularx}{\columnwidth}{@{}lcccc@{}}
        \toprule
        \textbf{Metrics}          & \textbf{400} & \textbf{800} & \textbf{1,200} & \textbf{12,000} \\
        \midrule  
        $C_D$ validation loss     & 0.0375       & 0.0335       & 0.0286         & 0.0266          \\
        Improvement               & -            & 10.67\%      & 23.73\%        & 29.07\%         \\
        \bottomrule
    \end{tabularx}
    \label{table:cdvalid}
\end{table}

\section{Conclusion}\label{sec:conclusion}

We present \dataset, a comprehensive automotive aerodynamics dataset comprising 12,000 high-fidelity \ac{cfd} simulations that bridges the gap between academic research and industrial applications. Generated using industry-standard \starccm software across three rear body configurations with 20 systematically varied \ac{cad} parameters, \dataset represents the most comprehensive and validated automotive aerodynamics dataset to date.

Our key contributions establish new standards for \ac{cfd} datasets in automotive engineering. We provide the highest-fidelity geometric representations available, incorporating complete engine compartments, cooling systems, and drivetrain components typically omitted in simplified academic models. The systematic parametric morphing using \ac{ffd} techniques creates geometric diversity while maintaining industrial relevance. Our rigorous experimental validation against wind tunnel measurements achieves exceptional accuracy with 1.04\% mean relative error for drag coefficients and excellent agreement in \ac{piv} flow field comparisons, establishing confidence in the dataset's predictive capability for real-world applications.

The industrial-grade simulation methodology employs production-level meshing protocols with 12 million hexahedral-dominant cells per case, capturing boundary layer phenomena and separation dynamics critical to automotive aerodynamics. Our comprehensive data formats provide complete access to flow physics information. The resulting 20TB dataset includes pressure coefficients, wall shear stress, velocity fields, streamlines, and iso-surfaces, enabling diverse research applications from fundamental flow physics to applied design optimization.

Our extensive machine learning benchmarks demonstrate that models trained on \dataset achieve industrial-grade prediction accuracy while reducing computational costs by orders of magnitude compared to traditional \ac{cfd} approaches. The systematic evaluation reveals fundamental trade-offs between geometric variability and prediction accuracy, providing crucial insights for developing robust aerodynamic prediction models and guiding future dataset development efforts.

\dataset's impact extends beyond immediate research applications to enable transformative changes in automotive development processes. By providing validated, high-fidelity training data, we enable the development of fast, accurate surrogate models that can replace computationally expensive \ac{cfd} simulations in design optimization loops. This supports rapid design exploration, real-time performance assessment, and integration of aerodynamic considerations throughout the vehicle development cycle.

The open availability of \dataset under permissive licensing, combined with comprehensive documentation and reproducible workflows, establishes a foundation for collaborative research in automotive aerodynamics. Future extensions will address current limitations, including multidisciplinary coupling with thermal management and structural dynamics, detailed in \cref{supp:sec:limitations}.

By democratizing access to industrial-quality aerodynamic data, \dataset empowers researchers worldwide to develop next-generation computational methods that accelerate the transition to more efficient, sustainable automotive designs. This work represents a critical step toward data-driven automotive engineering, where machine learning and high-fidelity simulation combine to revolutionize vehicle development practices.

\begin{ack}
    Deep learning experiments were conducted using the PaddlePaddle framework. We gratefully acknowledge Siemens Digital Industries Software for granting an educational-discount license of \starccm to Peking University, enabling part of the \ac{cfd} simulations presented in this study. Y. Xu, L. Cui, and Y. Zhu were supported by the National Natural Science Foundation of China (62376009), the PKU-BingJi Joint Laboratory for Artificial Intelligence, and the National Comprehensive Experimental Base for Governance of Intelligent Society, Wuhan East Lake High-Tech Development Zone.
\end{ack}

\bibliographystyle{apalike}

\clearpage

\section*{NeurIPS Paper Checklist}

\begin{enumerate}

\item {\bf Claims}
    \item[] Question: Do the main claims made in the abstract and introduction accurately reflect the paper's contributions and scope?
    \item[] Answer: \answerYes{} %
    \item[] Justification: Yes, the main claims made in the abstract and introduction accurately reflect our contribution and scope, which is the comprehensive, reproducible, and industrial-grade dataset \dataset. 

\item {\bf Limitations}
    \item[] Question: Does the paper discuss the limitations of the work performed by the authors?
    \item[] Answer: \answerYes{} %
    \item[] Justification: We have discussed the limitations in \cref{supp:sec:limitations}, including insufficient geometric deformation to cover all designs in actual industrial development, and the \ac{cfd} solver may still deviate from the real-industry conditions.
    \item[] Guidelines:
    \begin{itemize}
        \item The answer NA means that the paper has no limitation while the answer No means that the paper has limitations, but those are not discussed in the paper. 
        \item The authors are encouraged to create a separate "Limitations" section in their paper.
        \item The paper should point out any strong assumptions and how robust the results are to violations of these assumptions (\eg, independence assumptions, noiseless settings, model well-specification, asymptotic approximations only holding locally). The authors should reflect on how these assumptions might be violated in practice and what the implications would be.
        \item The authors should reflect on the scope of the claims made, \eg, if the approach was only tested on a few datasets or with a few runs. In general, empirical results often depend on implicit assumptions, which should be articulated.
        \item The authors should reflect on the factors that influence the performance of the approach. For example, a facial recognition algorithm may perform poorly when image resolution is low or images are taken in low lighting. Or a speech-to-text system might not be used reliably to provide closed captions for online lectures because it fails to handle technical jargon.
        \item The authors should discuss the computational efficiency of the proposed algorithms and how they scale with dataset size.
        \item If applicable, the authors should discuss possible limitations of their approach to address problems of privacy and fairness.
        \item While the authors might fear that complete honesty about limitations might be used by reviewers as grounds for rejection, a worse outcome might be that reviewers discover limitations that aren't acknowledged in the paper. The authors should use their best judgment and recognize that individual actions in favor of transparency play an important role in developing norms that preserve the integrity of the community. Reviewers will be specifically instructed to not penalize honesty concerning limitations.
    \end{itemize}

\item {\bf Theory assumptions and proofs}
    \item[] Question: For each theoretical result, does the paper provide the full set of assumptions and a complete (and correct) proof?
    \item[] Answer: \answerYes{} %
    \item[] Justification: We have provided a detailed explanation of the foundational theories of fluid dynamics and a thorough simulation setup clarification according to the theories, which can be found at \cref{supp:sec:turbulence_models}.

    \item {\bf Experimental result reproducibility}
    \item[] Question: Does the paper fully disclose all the information needed to reproduce the main experimental results of the paper to the extent that it affects the main claims and/or conclusions of the paper (regardless of whether the code and data are provided or not)?
    \item[] Answer: \answerYes{} %
    \item[] Justification: We provide the implementation details of the data generation pipeline at \cref{sec:dataset_pipeline,supp:sec:dataset_pipeline}.
    \item[] Guidelines:
    \begin{itemize}
        \item The answer NA means that the paper does not include experiments.
        \item If the paper includes experiments, a No answer to this question will not be perceived well by the reviewers: Making the paper reproducible is important, regardless of whether the code and data are provided or not.
        \item If the contribution is a dataset and/or model, the authors should describe the steps taken to make their results reproducible or verifiable. 
        \item Depending on the contribution, reproducibility can be accomplished in various ways. For example, if the contribution is a novel architecture, describing the architecture fully might suffice, or if the contribution is a specific model and empirical evaluation, it may be necessary to either make it possible for others to replicate the model with the same dataset, or provide access to the model. In general. releasing code and data is often one good way to accomplish this, but reproducibility can also be provided via detailed instructions for how to replicate the results, access to a hosted model (\eg, in the case of a large language model), releasing of a model checkpoint, or other means that are appropriate to the research performed.
        \item While NeurIPS does not require releasing code, the conference does require all submissions to provide some reasonable avenue for reproducibility, which may depend on the nature of the contribution. For example
        \begin{enumerate}
            \item If the contribution is primarily a new algorithm, the paper should make it clear how to reproduce that algorithm.
            \item If the contribution is primarily a new model architecture, the paper should describe the architecture clearly and fully.
            \item If the contribution is a new model (\eg, a large language model), then there should either be a way to access this model for reproducing the results or a way to reproduce the model (\eg, with an open-source dataset or instructions for how to construct the dataset).
            \item We recognize that reproducibility may be tricky in some cases, in which case authors are welcome to describe the particular way they provide for reproducibility. In the case of closed-source models, it may be that access to the model is limited in some way (\eg, to registered users), but it should be possible for other researchers to have some path to reproducing or verifying the results.
        \end{enumerate}
    \end{itemize}

\item {\bf Open access to data and code}
    \item[] Question: Does the paper provide open access to the data and code, with sufficient instructions to faithfully reproduce the main experimental results, as described in supplemental material?
    \item[] Answer: \answerYes{} %
    \item[] Justification: Both \dataset and code for data processing and benchmarking are accessible via our project \homepagelink{homepage}. We provide the detailed instructions for our dataset and code on the related websites.
    \item[] Guidelines:
    \begin{itemize}
        \item The answer NA means that paper does not include experiments requiring code.
        \item Please see the NeurIPS code and data submission guidelines (\url{https://nips.cc/public/guides/CodeSubmissionPolicy}) for more details.
        \item While we encourage the release of code and data, we understand that this might not be possible, so “No” is an acceptable answer. Papers cannot be rejected simply for not including code, unless this is central to the contribution (\eg, for a new open-source benchmark).
        \item The instructions should contain the exact command and environment needed to run to reproduce the results. See the NeurIPS code and data submission guidelines (\url{https://nips.cc/public/guides/CodeSubmissionPolicy}) for more details.
        \item The authors should provide instructions on data access and preparation, including how to access the raw data, preprocessed data, intermediate data, and generated data, etc.
        \item The authors should provide scripts to reproduce all experimental results for the new proposed method and baselines. If only a subset of experiments are reproducible, they should state which ones are omitted from the script and why.
        \item At submission time, to preserve anonymity, the authors should release anonymized versions (if applicable).
        \item Providing as much information as possible in supplemental material (appended to the paper) is recommended, but including URLs to data and code is permitted.
    \end{itemize}

\item {\bf Experimental setting/details}
    \item[] Question: Does the paper specify all the training and test details (\eg, data splits, hyperparameters, how they were chosen, type of optimizer, etc.) necessary to understand the results?
    \item[] Answer: \answerYes{} %
    \item[] Justification: We have made clear the dataset generation settings, reported how we validate and benchmark the SOTA models. More details can be found in the \cref{supp:sec:benchmark_results,supp:sec:exp_detail}.

\item {\bf Experiment statistical significance}
    \item[] Question: Does the paper report error bars suitably and correctly defined or other appropriate information about the statistical significance of the experiments?
    \item[] Answer: \answerNo{} %
    \item[] Justification: Due to computational resource constraints, we were unable to establish the statistical significance of our experimental results.
    \item[] Guidelines:
    \begin{itemize}
        \item The answer NA means that the paper does not include experiments.
        \item The authors should answer "Yes" if the results are accompanied by error bars, confidence intervals, or statistical significance tests, at least for the experiments that support the main claims of the paper.
        \item The factors of variability that the error bars are capturing should be clearly stated (for example, train/test split, initialization, random drawing of some parameter, or overall run with given experimental conditions).
        \item The method for calculating the error bars should be explained (closed form formula, call to a library function, bootstrap, etc.)
        \item The assumptions made should be given (\eg, Normally distributed errors).
        \item It should be clear whether the error bar is the standard deviation or the standard error of the mean.
        \item It is OK to report 1-sigma error bars, but one should state it. The authors should preferably report a 2-sigma error bar than state that they have a 96\% CI, if the hypothesis of Normality of errors is not verified.
        \item For asymmetric distributions, the authors should be careful not to show in tables or figures symmetric error bars that would yield results that are out of range (\eg negative error rates).
        \item If error bars are reported in tables or plots, The authors should explain in the text how they were calculated and reference the corresponding figures or tables in the text.
    \end{itemize}

\item {\bf Experiments compute resources}
    \item[] Question: For each experiment, does the paper provide sufficient information on the computer resources (type of compute workers, memory, time of execution) needed to reproduce the experiments?
    \item[] Answer: \answerYes{} %
    \item[] Justification: We have stated the high-performance computation cluster configuration and CPU core-hours needed in this paper to produce \dataset in \cref{supp:sec:compute_cost}.
    \item[] Guidelines:
    \begin{itemize}
        \item The answer NA means that the paper does not include experiments.
        \item The paper should indicate the type of compute workers CPU or GPU, internal cluster, or cloud provider, including relevant memory and storage.
        \item The paper should provide the amount of compute required for each of the individual experimental runs as well as estimate the total compute. 
        \item The paper should disclose whether the full research project required more compute than the experiments reported in the paper (\eg, preliminary or failed experiments that didn't make it into the paper). 
    \end{itemize}
    
\item {\bf Code of ethics}
    \item[] Question: Does the research conducted in the paper conform, in every respect, with the NeurIPS Code of Ethics \url{https://neurips.cc/public/EthicsGuidelines}?
    \item[] Answer: \answerYes{} %
    \item[] Justification: We have adhered to the NeurIPS Code of Ethics throughout the entire process of our research.
    \item[] Guidelines:
    \begin{itemize}
        \item The answer NA means that the authors have not reviewed the NeurIPS Code of Ethics.
        \item If the authors answer No, they should explain the special circumstances that require a deviation from the Code of Ethics.
        \item The authors should make sure to preserve anonymity (\eg, if there is a special consideration due to laws or regulations in their jurisdiction).
    \end{itemize}

\item {\bf Broader impacts}
    \item[] Question: Does the paper discuss both potential positive societal impacts and negative societal impacts of the work performed?
    \item[] Answer: \answerYes{} %
    \item[] Justification: We have stated that our work could bring positive societal impacts through boosting the automotive industry's development cycle and even extending beyond the field of automotive aerodynamics in \cref{supp:sec:impact}. 
    \item[] Guidelines:
    \begin{itemize}
        \item The answer NA means that there is no societal impact of the work performed.
        \item If the authors answer NA or No, they should explain why their work has no societal impact or why the paper does not address societal impact.
        \item Examples of negative societal impacts include potential malicious or unintended uses (\eg, disinformation, generating fake profiles, surveillance), fairness considerations (\eg, deployment of technologies that could make decisions that unfairly impact specific groups), privacy considerations, and security considerations.
        \item The conference expects that many papers will be foundational research and not tied to particular applications, let alone deployments. However, if there is a direct path to any negative applications, the authors should point it out. For example, it is legitimate to point out that an improvement in the quality of generative models could be used to generate deepfakes for disinformation. On the other hand, it is not needed to point out that a generic algorithm for optimizing neural networks could enable people to train models that generate Deepfakes faster.
        \item The authors should consider possible harms that could arise when the technology is being used as intended and functioning correctly, harms that could arise when the technology is being used as intended but gives incorrect results, and harms following from (intentional or unintentional) misuse of the technology.
        \item If there are negative societal impacts, the authors could also discuss possible mitigation strategies (\eg, gated release of models, providing defenses in addition to attacks, mechanisms for monitoring misuse, mechanisms to monitor how a system learns from feedback over time, improving the efficiency and accessibility of ML).
    \end{itemize}
    
\item {\bf Safeguards}
    \item[] Question: Does the paper describe safeguards that have been put in place for responsible release of data or models that have a high risk for misuse (\eg, pretrained language models, image generators, or scraped datasets)?
    \item[] Answer: \answerNA{} %
    \item[] Justification: Our paper poses no risk since we have only collected data from credible sources.

\item {\bf Licenses for existing assets}
    \item[] Question: Are the creators or original owners of assets (\eg, code, data, models), used in the paper, properly credited and are the license and terms of use explicitly mentioned and properly respected?
    \item[] Answer: \answerYes{} %
    \item[] Justification: We've credited the owner of \olddataset assets and stated the licenses of software we used (\starccm and \blender) in the paper.
    \item[] Guidelines:
    \begin{itemize}
        \item The answer NA means that the paper does not use existing assets.
        \item The authors should cite the original paper that produced the code package or dataset.
        \item The authors should state which version of the asset is used and, if possible, include a URL.
        \item The name of the license (\eg, CC-BY 4.0) should be included for each asset.
        \item For scraped data from a particular source (\eg, website), the copyright and terms of service of that source should be provided.
        \item If assets are released, the license, copyright information, and terms of use in the package should be provided. For popular datasets, \url{paperswithcode.com/datasets} has curated licenses for some datasets. Their licensing guide can help determine the license of a dataset.
        \item For existing datasets that are re-packaged, both the original license and the license of the derived asset (if it has changed) should be provided.
        \item If this information is not available online, the authors are encouraged to reach out to the asset's creators.
    \end{itemize}

\item {\bf New assets}
    \item[] Question: Are new assets introduced in the paper well documented and is the documentation provided alongside the assets?
    \item[] Answer: \answerYes{} %
    \item[] Justification: Both \dataset and code for data processing and benchmarking are well documented, and are accessible via our project \homepagelink{homepage}. 

\item {\bf Crowdsourcing and research with human subjects}
    \item[] Question: For crowdsourcing experiments and research with human subjects, does the paper include the full text of instructions given to participants and screenshots, if applicable, as well as details about compensation (if any)? 
    \item[] Answer: \answerNA{} %
    \item[] Justification: This paper does not involve crowdsourcing nor research with human subjects.

\item {\bf Institutional review board (IRB) approvals or equivalent for research with human subjects}
    \item[] Question: Does the paper describe potential risks incurred by study participants, whether such risks were disclosed to the subjects, and whether Institutional Review Board (IRB) approvals (or an equivalent approval/review based on the requirements of your country or institution) were obtained?
    \item[] Answer: \answerNA{} %
    \item[] Justification: This paper does not involve crowdsourcing nor research with human subjects.

\item {\bf Declaration of LLM usage}
    \item[] Question: Does the paper describe the usage of LLMs if it is an important, original, or non-standard component of the core methods in this research? Note that if the LLM is used only for writing, editing, or formatting purposes and does not impact the core methodology, scientific rigorousness, or originality of the research, declaration is not required.
    \item[] Answer: \answerNA{} %
    \item[] Justification: LLMs are not involved in our core method development in this research.
\end{enumerate}

\clearpage
\appendix
\renewcommand\thefigure{A\arabic{figure}}
\setcounter{figure}{0}
\renewcommand\thetable{A\arabic{table}}
\setcounter{table}{0}
\renewcommand\theequation{A\arabic{equation}}
\setcounter{equation}{0}
\pagenumbering{arabic}%
\renewcommand*{\thepage}{A\arabic{page}}
\setcounter{footnote}{0}

\section{Resources Availability and Licensing}\label{supp:sec:licensing}

\dataset is distributed under the \ac{license} license to facilitate academic dissemination while protecting intellectual property rights. Under this license, users are permitted to copy, distribute, display, and perform the work, as well as create derivative works, provided that they give appropriate credit to the original authors, provide a link to the license, and indicate if changes were made. \textbf{Commercial use of the work is strictly prohibited without explicit written permission from the copyright holders.}

All associated resources, including source code, datasets, and documentation, are publicly accessible to ensure transparency and enable reproducibility. Users are encouraged to contribute improvements through established channels, enhancing the collective value of these open-source resources within the academic community.

\section{Broader Impact}\label{supp:sec:impact}

\dataset addresses critical needs across multiple research and industrial domains. In academic research, the high-fidelity dataset enables development of novel geometric deep learning models for complex fluid dynamics, advancing understanding of how neural networks capture physical phenomena, and supporting investigations into physics-informed machine learning approaches. The dataset expands available benchmarks for graph neural networks and point cloud methods, establishing new evaluation standards for fluid dynamics applications.

For industrial applications, \dataset enables development of fast surrogate models that can replace computationally expensive \ac{cfd} simulations in design optimization workflows. This capability supports more efficient automotive development cycles and extends to aerospace applications where virtual testing reduces reliance on expensive physical prototypes. The dataset addresses growing industrial demand for understanding deep learning generalization across large solution domains, promoting integration of high-fidelity physics-based simulations with artificial intelligence.

Beyond immediate applications, \dataset establishes a foundation for data-driven design methodologies that can transform engineering practices across disciplines facing computational constraints. By democratizing access to industrial-quality aerodynamic data, the dataset empowers researchers worldwide to develop next-generation computational methods for more efficient and sustainable engineering designs.

\begin{table}[b!]
    \centering
    \small
    \caption{\textbf{Parametric deformation ranges for automotive geometry generation via lattice-based morphing.} Twenty parameters control geometric variations across vehicle components, with ranges established from production vehicle design standards and aerodynamic optimization requirements.}
    \label{tab:supp:car-parameters}
    \begin{tabular}{lllll}
        \toprule
        \textbf{Number} & \textbf{Parameter} & \textbf{Minimum} & \textbf{Maximum} & \textbf{Unit} \\
        \midrule
        {1} & Global Scaling & $80\%$ & $120\%$ & - \\
        {2} & Vehicle Width & -0.1 & 0.1 & m \\
        {3} & Vehicle Length & -0.1 & 0.1 & m \\
        {4} & Ramp Angle & -10 & 10 & degree \\
        {5} & Front Bumper Length & -0.1 & 0.1 & m \\
        {6} & Windscreen X & -0.05 & 0.05 & m \\
        {7} & Windscreen Z & -0.05 & 0.05 & m \\
        {8} & Side Mirrors X & -0.05 & 0.05 & m \\
        {9} & Side Mirrors Z & -0.05 & 0.05 & m \\
        {10} & Rear Window X & -0.05 & 0.05 & m \\
        {11} & Rear Window Z & -0.05 & 0.05 & m \\
        {12} & Trunk Lid Angle & -10 & 10 & degree \\
        {13} & Trunk Lid X & -0.05 & 0.05 & m \\
        {14} & Trunk Lid Z & -0.05 & 0.05 & m \\
        {15} & Diffuser Angle & -10 & 10 & degree \\
        {16} & Greenhouse Angle & -10 & 10 & degree \\
        {17} & Front Hood Angle & -10 & 10 & degree \\
        {18} & Intake Hood Angle & -10 & 10 & degree \\
        {19} & Tire Diameter & -0.033 & 0.033 & m \\
        {20} & Tire Width & -0.015 & 0.015 & m \\
        \bottomrule
    \end{tabular}
\end{table}

\section{Dataset Generation, Simulation, and Validation}\label{supp:sec:dataset_pipeline}

\subsection{Geometric Morphing via Reference Vehicle Architecture and Parametric Design}\label{supp:sec:ffd}

\begin{figure}[t!]
    \centering
    \includegraphics[width=\linewidth]{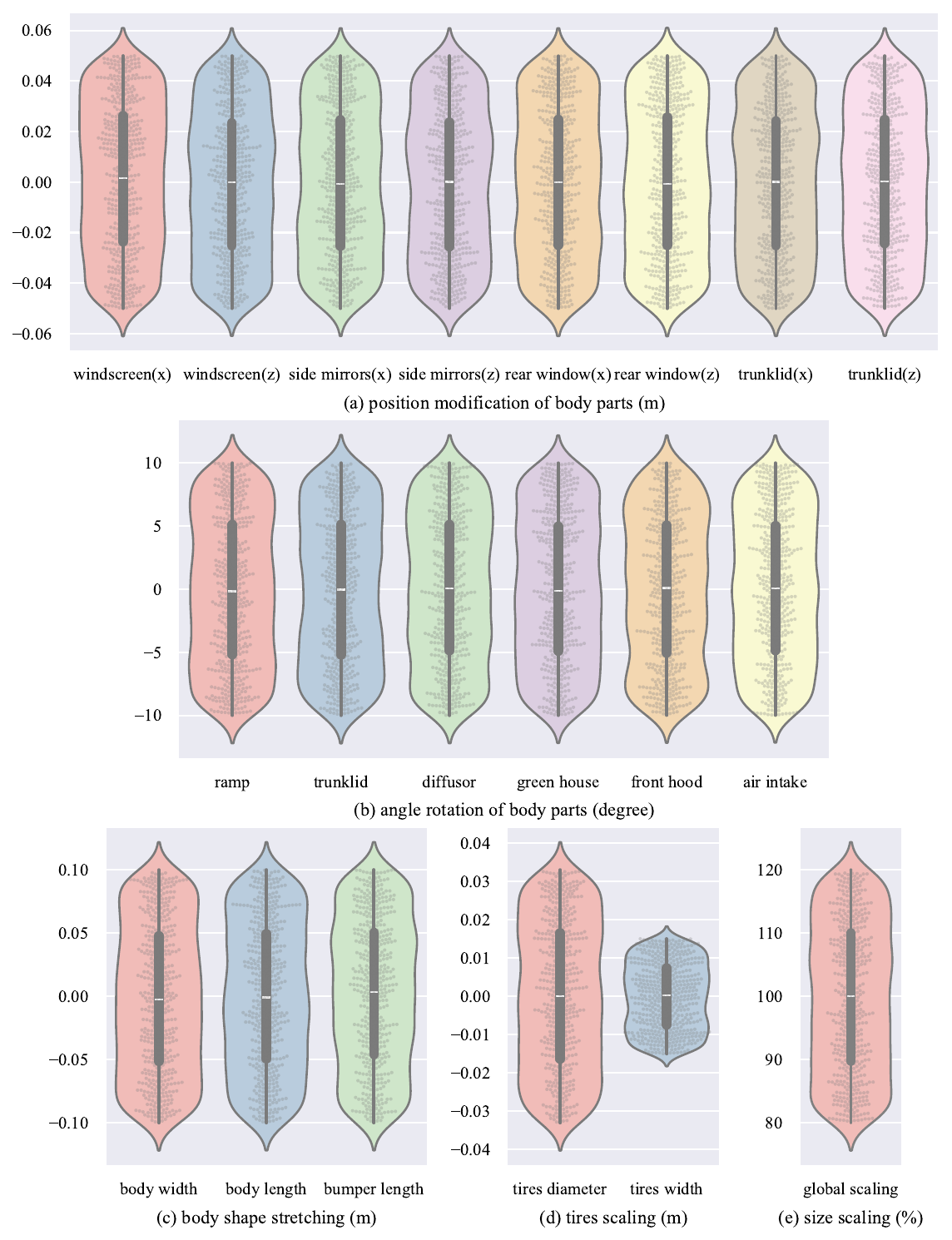} 
    \caption{\textbf{Parametric deformation diversity in \dataset.} Five deformation categories generate geometric variations through Latin Hypercube Sampling: (a) Linear displacement of four vehicle components within ±0.05m in x and z directions. (b) Angular rotation of six body elements ranging from $-10\degree$ to $10\degree$. (c) Stretching and compression of vehicle body and front bumper between ±0.1m. (d) Tire deformation with diameter variations of ±0.033m and width variations of ±0.015m. (e) Global scaling from 80\% to 120\% of baseline dimensions.}
    \label{supp:fig:violin} 
\end{figure}

We develop a parametric geometry generation framework using lattice deformation to create diverse automotive geometries from the open-source \olddataset vehicle model. Our approach employs \ac{ffd} techniques implemented through Python scripts on the \blender platform, establishing a $32 \times 8 \times 8$ three-dimensional lattice structure that enables fine-grained control over local body regions while maintaining geometric integrity.

The implementation begins with bounding box analysis to determine spatial distribution of geometric features, establishing mapping relationships between local and global coordinate systems. The parametric framework divides the vehicle into nine key regions including front bumper, trunk, and roof, with regional deformation control achieved through predefined index grouping. We define 20 key parameters covering body morphology, dimensional scaling, and wheel configurations (\cref{tab:supp:car-parameters}). Parameter ranges reference production vehicle design standards and incorporate \ac{cfd} simulation experience to set boundary conditions that ensure engineering feasibility, with diffuser inclination constrained within $\pm 10^\circ$ to balance downforce and drag requirements.

\begin{figure}[b!]
    \centering
    \includegraphics[width=\linewidth]{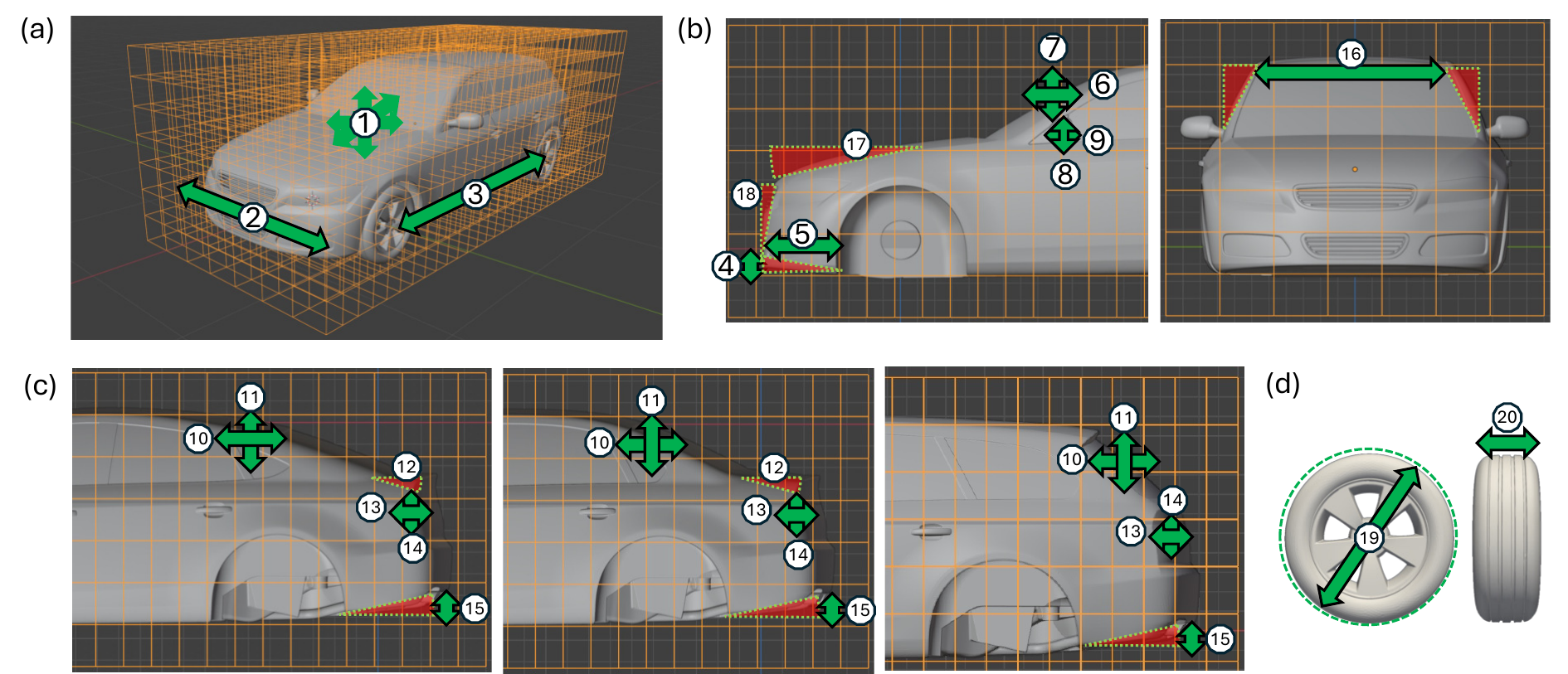}
    \caption{\textbf{Lattice deformation framework for \olddataset vehicle configurations.} (a) Deformation lattice structure with vehicle scaling, width, and length adjustment parameters. (b) Front vehicle section parameters controlling bumper and forward geometry modifications. (c) Rear section parameter sets for three vehicle types: fastback (left), notchback (center), and estateback (right) configurations. (d) Wheel geometry parameters for tire diameter and width variations.}
    \label{supp:fig:deform}
\end{figure}

\begin{figure}[t!]
    \centering
    \includegraphics[width=\linewidth]{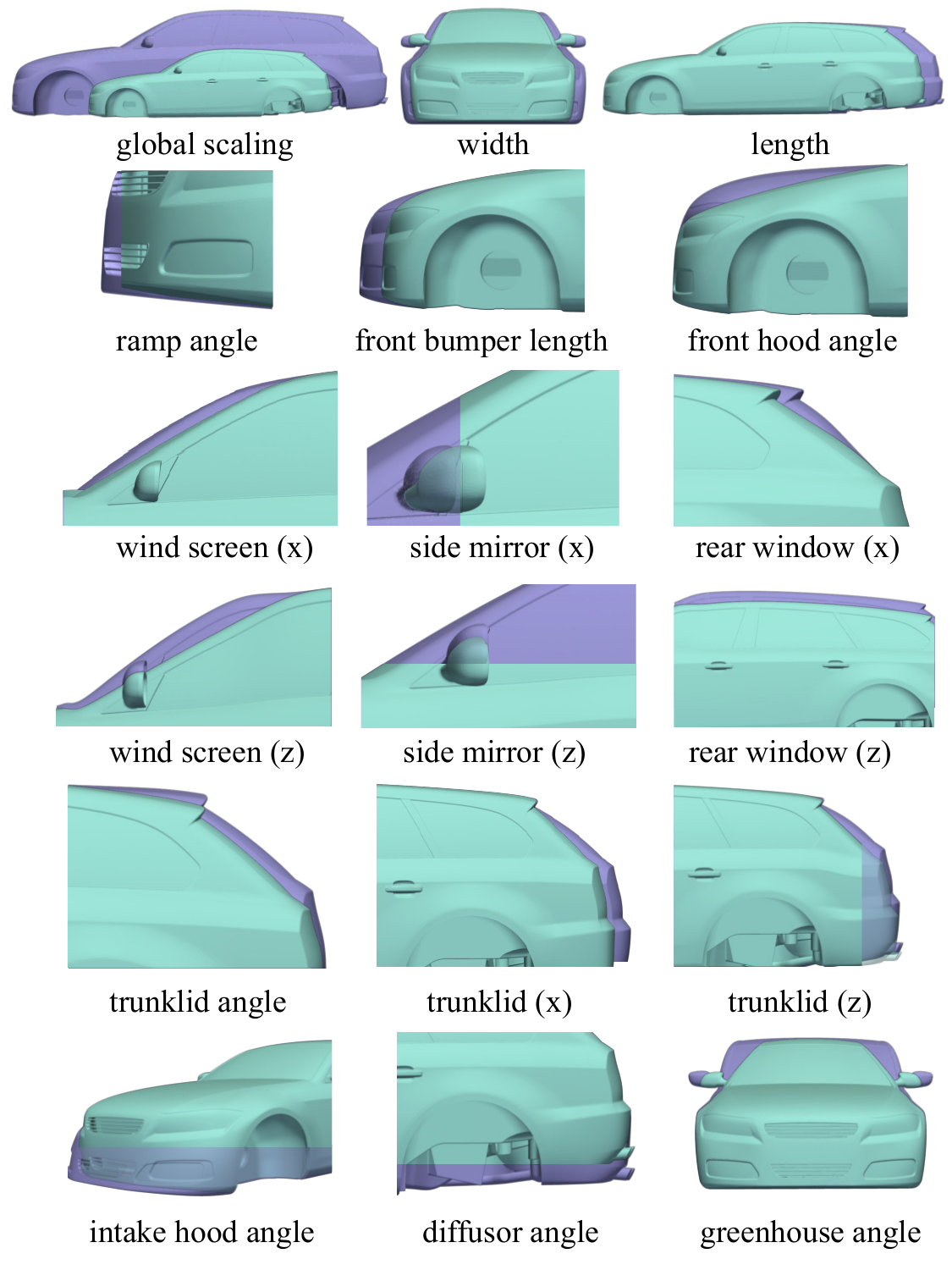} 
    \caption{\textbf{Individual parameter deformation visualization.} Lattice deformation results for single-parameter variations across minimum to maximum ranges. Blue models represent minimum parameter values, purple models represent maximum values. Wheel size parameters are excluded to emphasize vehicle body geometry modifications.}
    \label{supp:fig:deform_detail} 
\end{figure}

\begin{figure}[t!]
    \centering
    \includegraphics[width=\linewidth]{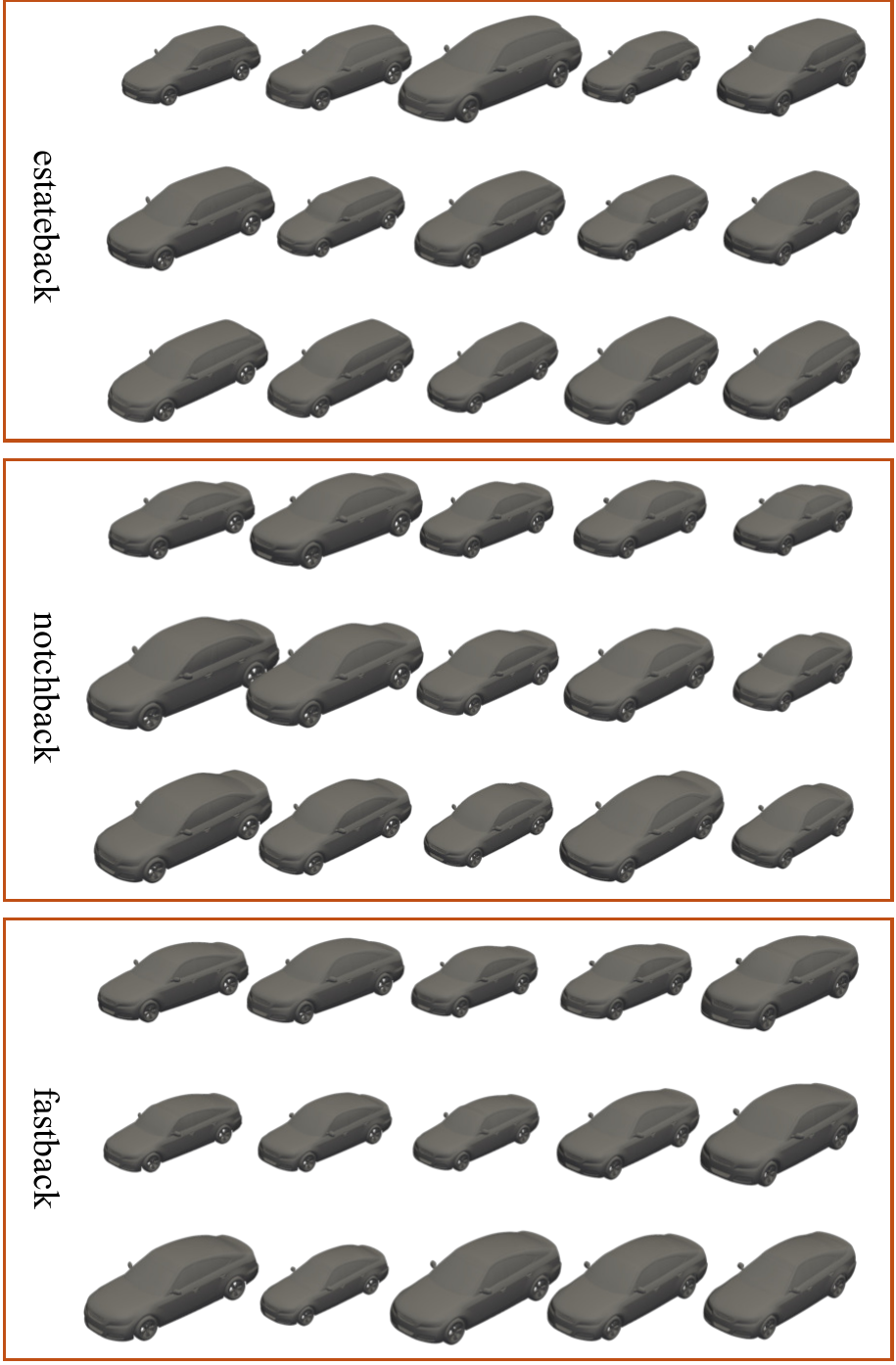} 
    \caption{\textbf{Multi-parameter deformation examples by vehicle type.} Deformation variations for three rear configurations: estateback (top), notchback (middle), and fastback (bottom).}
    \label{supp:fig:shapes_park} 
\end{figure}

\ac{lhs} generates 1000 parameter samples across engineering-reasonable ranges for statistical representativeness, producing thousands of unique geometric configurations. Each parameter corresponds to specific physical modifications: dimensional stretches (front bumper length $\pm 0.1$m), positional adjustments (trunk inclination $\pm 0.05$m), size scaling (vehicle width $\pm 0.1$m), and angular rotations (windscreen inclination $\pm 0.05$m). Parameter distributions are detailed in \cref{supp:fig:violin}, with each configuration generating unique variants through nine independent deformation functions.

For cooling system integration---absent in previous datasets---we bind front grille, radiator, engine bay, and gearbox components to unified lattice deformers using \ac{lhs} methodology. This coordinated deformation prevents spatial interpenetration while maintaining positional coherence among complex internal components during geometric transformations.

Precise wheel alignment employs a dedicated four-wheel positioning algorithm with predefined feature points including 3D coordinate systems for all four wheels. The algorithm dynamically calculates axle positions by tracking spatial transformations of these feature points during deformation, ensuring geometric matching between axles and deformed body while maintaining lattice deformation controllability and avoiding component distortion issues common in traditional morphing techniques.

This lattice deformation approach reduces geometric generation time by over 90\% compared to traditional \ac{cad} parametric methods while preserving industrial-grade surface quality. The systematic parameterization via \ac{ffd} maintains geometric continuity, topological consistency, and engineering feasibility while establishing a deformable configuration space rooted in industrial design practices. As illustrated in \cref{supp:fig:deform,supp:fig:deform_detail,supp:fig:shapes_park}, the method preserves aerodynamic baseline characteristics of the reference vehicle while enabling parametric integration of engineering details such as cooling systems and chassis components, generating complex geometric interactions including length-width coupling variations that provide foundation for neural networks to capture nonlinear flow features.

Future research will integrate \ac{rans} simulation results to establish high-dimensional mapping models between morphological parameters and aerodynamic coefficients, supporting development of intelligent aerodynamic optimization frameworks.

\subsection{Surface Wrapping Implementation Details}\label{supp:sec:surface_wrap}

\begin{figure}[ht!]
    \centering
    \begin{subfigure}{0.333\linewidth}
        \includegraphics[width=\linewidth]{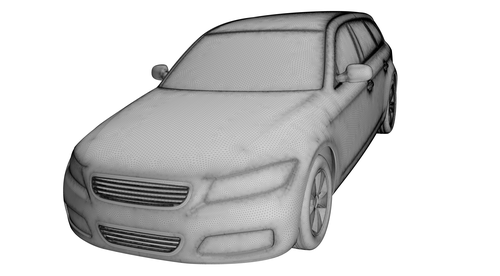}
        \caption{initial surface mesh}
    \end{subfigure}%
    \begin{subfigure}{0.333\linewidth}
        \includegraphics[width=\linewidth]{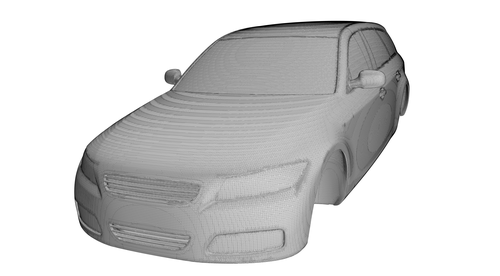}
        \caption{wrapped vehicle body}
    \end{subfigure}%
    \begin{subfigure}{0.333\linewidth}
        \includegraphics[width=\linewidth]{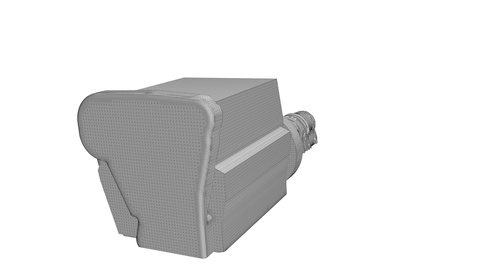}
        \caption{wrapped engine and gearbox}
    \end{subfigure}%
    \\
    \begin{subfigure}{0.333\linewidth}
        \includegraphics[width=\linewidth]{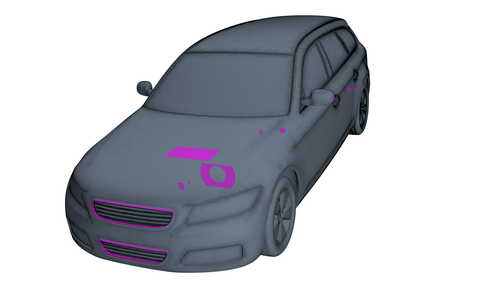}
        \caption{close proximity faces}
    \end{subfigure}%
    \begin{subfigure}{0.333\linewidth}
        \includegraphics[width=\linewidth]{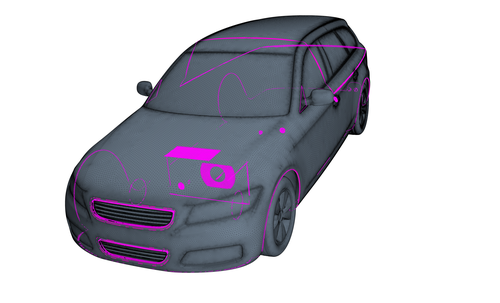}
        \caption{pierced faces}
    \end{subfigure}%
    \begin{subfigure}{0.333\linewidth}
        \includegraphics[width=\linewidth]{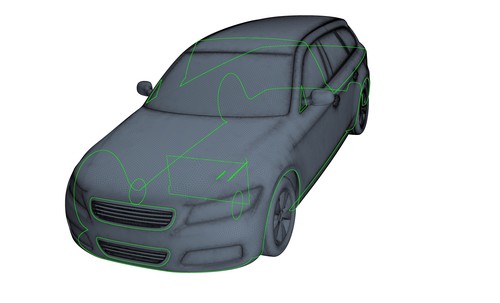}
        \caption{free edges}
    \end{subfigure}%
    \caption{\textbf{Surface wrapping process using \starccm.} Surface wrapping workflow applied to vehicle components: (a) Initial generated surface mesh with critical defects. (b) Wrapped vehicle body mesh. (c) Wrapped engine and gearbox components. The process addresses three fundamental surface errors: (d) close proximity faces, (e) pierced faces, and (f) free edges, with defective faces highlighted in purple and problematic edges in green.}
    \label{supp:fig:surface_wrapping}
\end{figure}

Generated vehicle geometries exhibit critical surface defects that prevent direct \ac{cfd} computation, requiring mandatory surface wrapping preprocessing for all components. As illustrated in \cref{supp:fig:surface_wrapping}, we employ \starccm's surface wrapping functionality to independently process each component (body, drivetrain, wheels) of the initial surface mesh, addressing three fundamental mesh errors: (i) \textit{Close Proximity} faces with minimum distances below 0.1 mm, (ii) \textit{Pierced Faces} that create flow leakage paths, and (iii) \textit{Free Edges} that compromise watertight geometric continuity.

Aerodynamically critical regions, including grilles and wheel arches, receive enhanced treatment through local refinement with 10\% size reduction and curvature adaptation using a minimum spacing of 0.5 mm. Post-wrapping validation ensures aspect ratios remain below 20 and skewness angles exceed 10° across all mesh components. This optimized workflow achieves 98.6\% defect reduction while maintaining 0.28 mm geometric fidelity, enabling reliable \ac{cfd} simulations with aerodynamic feature resolution below 1 mm accuracy.

\subsection{Regional Mesh}\label{supp:sec:mesh}

\paragraph{Mesh Generation Framework}

The parallel mesh generation employs a global base size $L_0 = 0.24\,\mathrm{m}$ with surface growth rate 1.3, utilizing hexahedral elements enhanced by triangular surface reconstruction for geometric fidelity. Dual refinement strategies ensure critical feature capture: curvature refinement with $0.01\,\mathrm{m}$ deviation distance and 36 points per circle, plus proximity refinement for chassis, engine bay, and wheel components.

Volume growth adopts \textit{very slow} mode with maximum element size constrained to $0.48\,\mathrm{m}$ ($200\% L_0$) and minimum surface size to $0.024\,\mathrm{m}$ ($10\% L_0$). Three-tiered volume controls implement progressive refinement: Block 2 at $0.06\,\mathrm{m}$ ($25\% L_0$), Block 3 matching $L_0$, and Block 4 at $0.024\,\mathrm{m}$ ($10\% L_0$).

\paragraph{Surface and Boundary Layer Treatment}

Vehicle body surfaces receive enhanced treatment with target surface size $0.0144\,\mathrm{m}$ ($6\% L_0$) and minimum size $0.0096\,\mathrm{m}$ ($4\% L_0$). Prism layers follow geometric progression with first-layer height $h_1 = 0.005\,\mathrm{mm}$, expansion ratio $r = 1.2$, and 8-layer configuration yielding total thickness $H = h_1 \cdot (r^8 - 1)/(r - 1) \approx 0.114\,\mathrm{mm}$. External boundaries (inlet/outlet/side/top) utilize coarser $0.48\,\mathrm{m}$ elements without prism layer refinement.

\paragraph{Quality Control and Validation}

Mesh quality maintains triangular surface elements above a 0.05 minimum threshold with boundary layer resolution enforcing $y^+ > 30$ using blended wall functions. Mesh independence verification employs three resolutions---coarse ($0.48\,\mathrm{m}$), medium ($0.24\,\mathrm{m}$), and fine ($0.12\,\mathrm{m}$)---validated through drag coefficient and flow field analysis.

\subsection{Navier–Stokes Equations and Turbulence Model}\label{supp:sec:turbulence_models}

\paragraph{Reynolds-Averaged Navier-Stokes Approach}

Automotive external aerodynamics simulation requires appropriate turbulent flow treatment. While \ac{dns} captures all turbulence scales, it demands prohibitive computational resources for practical automotive applications. This study employs the \ac{rans} approach, modeling time-averaged flow fields at significantly reduced computational cost compared to \ac{dns} or \ac{les} alternatives.

The \ac{rans} equations derive from Reynolds decomposition, where each flow variable decomposes into mean and fluctuating components: $\phi=\bar{\phi}+\phi'$. For steady-state incompressible flow, the governing equations become:
\begin{align}    
    \frac{\partial \rho}{\partial t}+\nabla \cdot(\rho \bar{\mathbf{v}})&=0, \\
    \frac{\partial}{\partial t}(\rho \bar{\mathbf{v}})+\nabla \cdot(\rho \bar{\mathbf{v}} \otimes \bar{\mathbf{v}})&=-\nabla \bar{p} +\nabla \cdot\left(\bar{\mathbf{T}}+\mathbf{T}_{\text{RANS}}\right), 
\end{align}
where $\rho$, $\bar{\mathbf{v}}$, $\bar{p}$, and $\bar{\mathbf{T}}$ represent air density, mean velocity vector, mean pressure, and mean viscous stress tensor, respectively. The Reynolds stress tensor $\mathbf{T}_{\text{RANS}}$ requires turbulence modeling closure:
\begin{equation}
    \mathbf{T}_{\text{RANS}} = -\rho \begin{pmatrix}
    \overline{u'u'} & \overline{u'v'} & \overline{u'w'} \\
    \overline{u'v'} & \overline{v'v'} & \overline{v'w'} \\
    \overline{u'w'} & \overline{v'w'} & \overline{w'w'}
    \end{pmatrix},
\end{equation}
where $u'$, $v'$, $w'$ represent velocity fluctuations in the $x$-, $y$-, $z$-directions, respectively.

\paragraph{SST $k$-$\omega$ Turbulence Model}

This study employs the \ac{sst} $k$-$\omega$ turbulence model~\citep{menter1994two}, solving transport equations for turbulent kinetic energy $k$ and specific dissipation rate $\omega$. This model provides superior performance for boundary layers under adverse pressure gradients and applies throughout the boundary layer without wall-distance computation, yielding a reliable approximation for automotive aerodynamics.

The transport equations are formulated as:
\begin{align}
    \frac{\partial}{\partial t}(\rho k)+\nabla \cdot(\rho k \bar{\mathbf{v}})&=\nabla \cdot\left[\left(\mu+\sigma_{k} \mu_{t}\right) \nabla k\right]+P_{k}-\rho \beta^{*} f_{\beta^{*}}\left(\omega k-\omega_{0} k_{0}\right), \\
    \frac{\partial}{\partial t}(\rho \omega)+\nabla \cdot(\rho \omega \bar{\mathbf{v}})&=\nabla \cdot\left[\left(\mu+\sigma_{\omega} \mu_{t}\right) \nabla \omega\right]+P_{\omega}-\rho \beta f_{\beta}\left(\omega^{2}-\omega_{0}^{2}\right),\\
    \mu_{t}&= \min\left(\frac{1}{\omega}, \frac{1}{F_2S}\right),\\
    \phi&=F_1 \phi_1 +(1-F_1)\phi_2, \quad (\phi=\sigma_k,\sigma_\omega,\beta),
\end{align}
where $\mu$ and $\mu_t$ denote dynamic and turbulent eddy viscosity, $f_{\beta*}$ and $f_\beta$ represent free-shear and vortex-stretching modification factors, $k_0$ and $\omega_0$ are ambient turbulence values~\citep{spalart2007effective}, and $F_1$, $F_2$ are blending functions connecting inner and outer boundary layer regions.

Model parameters are: $\sigma_{k1}=0.85$, $\sigma_{k2}=1.0$, $\sigma_{\omega1}=0.5$, $\sigma_{\omega2}=0.856$, $\beta_1=0.075$, $\beta_2=0.0828$. Following the Boussinesq hypothesis, the Reynolds stress tensor is modeled as:
\begin{align}
    \mathbf{T}_{\text{RANS}}=\mu_{t}\left(\frac{\partial \bar{u}_{i}}{\partial x_{j}} + \frac{\partial \bar{u}_{j}}{\partial x_{i}}\right) - \frac{2}{3}\rho k \delta_{ij},
\end{align}   
providing complete system closure.

\paragraph{Porous Media Treatment}

Vehicle cooling components (radiators, condensers) are modeled as porous media due to their complex internal geometries that are too fine for individual meshing. The void fraction $\varepsilon_{g}$ represents the geometry at coarse mesh scales, while momentum exchange is captured through body force terms.

The governing equations for porous media become:
\begin{align}
    \frac{\partial(\varepsilon_{g}\rho)}{\partial t}+\nabla\cdot(\rho\bar{\mathbf{v}}) &= 0, \\
    \frac{\partial(\rho\bar{\mathbf{v}})}{\partial t}+\nabla\cdot(\rho\bar{\mathbf{v}}\otimes\bar{\mathbf{v}})&=-\nabla p+\nabla\cdot(\bar{\mathbf{T}}+\mathbf{T}_{\text{RANS}})+\mathbf{F}_{p},
\end{align}
where the flow resistance $\mathbf{F}_{p}$ is given by:
\begin{equation}
    \mathbf{F}_{p}=-\left(\mathbf{P}_{v}+\mathbf{P}_{i}|\bar{\mathbf{v}}|\right)\cdot\bar{\mathbf{v}},
\end{equation}
with $\mathbf{P}_{v}$ and $\mathbf{P}_{i}$ representing viscous and inertial resistance tensors. The superficial velocity $\bar{\mathbf{v}}$ relates to the actual pore velocity $\mathbf{v}_{g}$ through $\bar{\mathbf{v}}=\varepsilon_{g}\mathbf{v}_{g}$.

Further implementation details are available in the \href{https://docs.sw.siemens.com/documentation/external/PL20200805113346338/en-US/userManual/userguide/html/index.html#page/STARCCMP%2FGUID-235E939A-BC77-4988-AE0A-D79B17FD6072.html%23}{\starccm Documentation}.

\subsection{Computational Infrastructure}\label{supp:sec:compute_cost}

Dataset generation utilized a high-performance computing cluster with 100 nodes, each equipped with Intel\textregistered\ Xeon\textregistered\ Gold 6148 processors, consuming approximately 1,080,000 core-hours. Computational efficiency was optimized through the \ac{amg} linear solver~\citep{stuben2001review}, which accelerates convergence while reducing memory requirements.

\subsection{Simulation Result}\label{supp:sec:sim_result}

\begin{figure}[t!]
    \centering
    \begin{subfigure}{0.634\linewidth}
        \includegraphics[width=\linewidth]{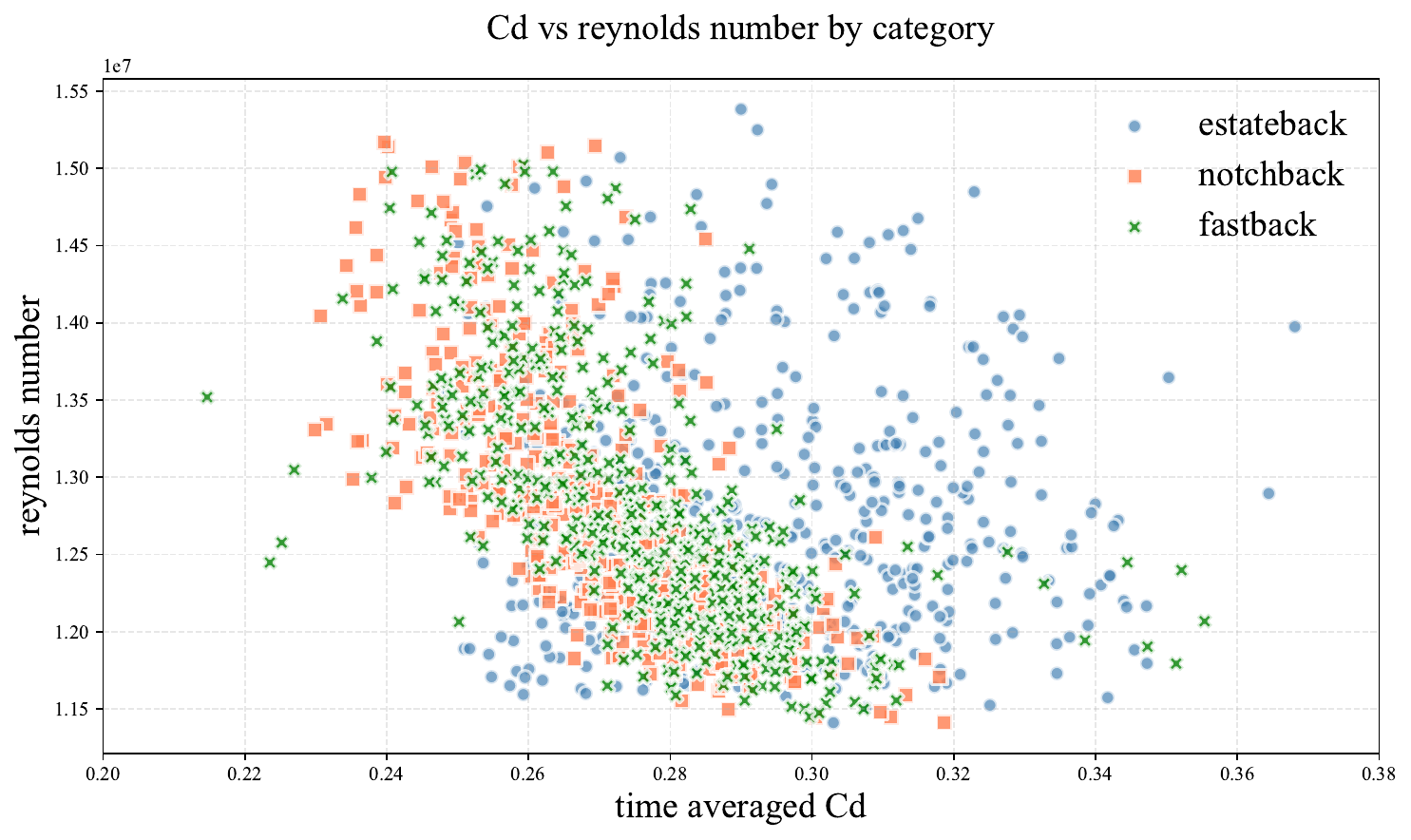}
        \caption{ }
    \end{subfigure}%
    \begin{subfigure}{0.366\linewidth}
        \includegraphics[width=\linewidth]{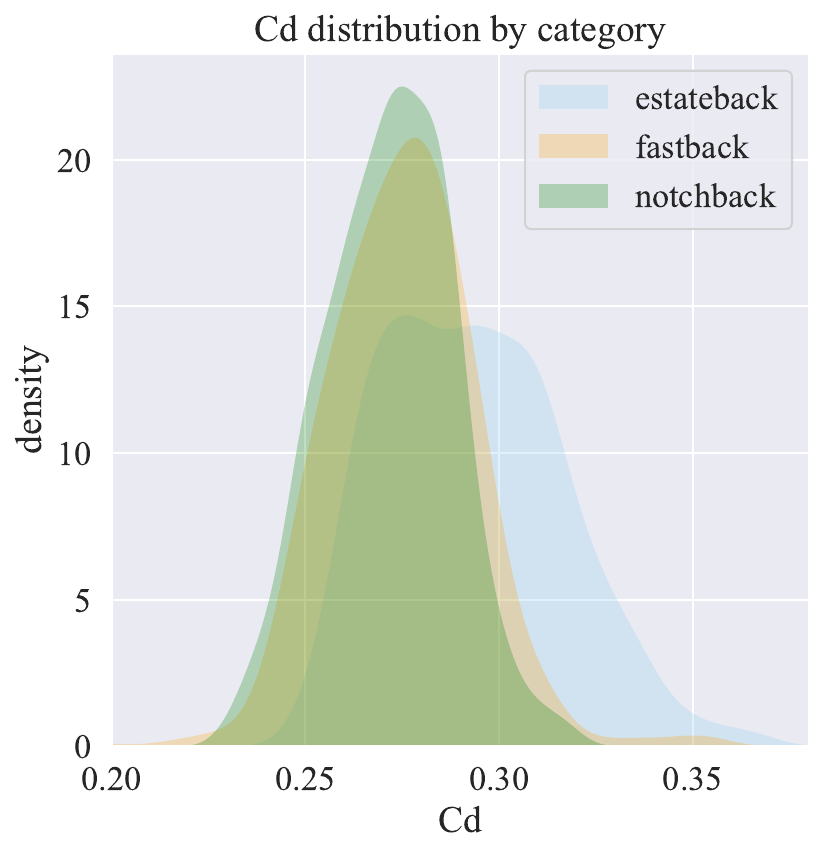}
        \caption{ }
    \end{subfigure}%
    \caption{\textbf{Aerodynamic performance comparison of three rear configurations.} (a) Drag coefficient ($C_D$) versus Reynolds number ($Re$) with 95\% confidence intervals (shaded regions). (b) Probability density distributions of $C_D$ values highlighting modal characteristics. Estateback (blue), notchback (orange), and fastback (green) configurations are consistently color-coded across subfigures.}
    \label{supp:fig:cd_comparison}
\end{figure}

The aerodynamic behavior of three rear configurations (estateback, notchback, fastback) is quantified in \cref{supp:fig:cd_comparison}, with comprehensive flow field visualizations including pressure distributions (\cref{supp:fig:pressure_park}), wall shear stress patterns (\cref{supp:fig:wss_park}), and velocity profiles (\cref{supp:fig:velocity_park}).

\paragraph{Reynolds Number Dependencies}

Drag characteristics exhibit distinct Reynolds number trends across configurations. The notchback configuration demonstrates consistent $C_D$ reduction with increasing $Re$ throughout the observed range, while the fastback shows progressive $C_D$ decrease across its operational regime. In contrast, the estateback maintains relatively uniform $C_D$ values without significant monotonic variation, though with broader dispersion compared to other configurations.

\paragraph{Statistical Characteristics}

The estateback exhibits bimodal density peaks at $C_D = 0.31$ and $0.35$, contrasting with unimodal distributions of notchback ($\mu = 0.28$, IQR = 0.04) and fastback ($\mu = 0.33$). The notchback's narrow interquartile range confirms superior aerodynamic consistency, while the fastback's right-skewed distribution reflects intermittent high-drag flow states.

\subsection{Computational Validation and Quality Assurance}\label{supp:sec:validation_quality}

\paragraph{Automated Quality Control}

A comprehensive quality assessment framework with specific acceptance criteria ensures dataset integrity. Automated filtering verifies that physical quantities remain within engineering ranges: pressure values (volumetric and surface) and wall shear stress do not exceed $\pm 2 \times 10^4$ Pa. Simulations exhibiting numerical anomalies are automatically excluded from the dataset.

\paragraph{Convergence Criteria}

All cases undergo minimum 1,000 iteration steps with convergence determined by $C_D$ stabilization within $\pm 1\%$ over the final 100 iterations. A hybrid convergence metric considers both relative $C_D$ change and asymptotic behavior of solution residuals across continuity, turbulence, and momentum equations. Cases with unresolved oscillatory behavior or non-physical trends are identified through spectral analysis and excluded.

\paragraph{Mesh and Solution Quality}

Mesh quality verification ensures appropriate wall treatment application, with all simulations maintaining wall $y^+$ values outside the buffer layer for accurate near-wall flow resolution using blended wall functions. Meshes with significant skewness or aspect ratio defects undergo refinement or exclusion. Final quality control discards simulations with the lowest 5\% composite quality scores, retaining only the top 60\% of simulations based on computational quality metrics for subsequent analysis.

\subsection{Differences Between Simulation and Experiments}\label{supp:sec:diffb}

\paragraph{Geometric Configuration Differences}

The primary geometric differences between simulation and experimental setups concentrate on wheel mounting systems and radiator representation. The front wheels utilize MacPherson strut-type mounting assemblies, while the rear wheels employ solid axle configurations. The experimental radiator geometry is unavailable in the \ac{cad} model and is instead represented through porous media modeling with pressure drop characteristics derived from hexagonal aluminum honeycomb and perforated sheet specifications.

The experimental model features five-spoke wheels with detailed tire tread patterns, symmetric mirror assemblies, comprehensive chassis geometry, and three rear configurations: fastback, notchback, and estateback. A precisely machined floor section maintains 4 mm wheel clearances, while detailed engine bay components include integrated cooling systems and spotlights. Wheel assembly tolerances are maintained within $\pm 0.5$ mm.

\paragraph{Physical Modeling Validation}

\cref{table:cdvalidwind} presents drag coefficient comparison between experimental measurements and \dataset simulation results. All three rear configuration coefficients have been calibrated to approximate open-road operating conditions. The primary error sources include: flow regime complexities around the four-wheel assembly, pressure drop modeling through radiator porosity representation, and inherent turbulence model limitations in capturing separation and reattachment phenomena.

The validation demonstrates acceptable agreement between computational and experimental results, with discrepancies primarily attributed to geometric simplifications in the cooling system representation and the challenge of accurately modeling complex wheel-ground interactions in the computational domain.

\section{Benchmark Setup and Evaluation}

\subsection{Evaluation Metrics}\label{supp:sec:eval_metrics}

\paragraph{Relative $L_2$ Loss ($\epsilon$)}

The relative $L_2$ error serves as the primary evaluation metric for flow field regression across all experimental configurations. To ensure comparability between different physical variables, each flow field variable $\phi$ (velocity, pressure, wall shear stress) undergoes standardization using dataset-wide statistics:
\begin{equation}
    \hat{\phi} = \frac{\phi - \mu_\phi}{\sigma_\phi},
\end{equation}
where $\mu_\phi$ and $\sigma_\phi$ represent the mean and standard deviation computed over the entire dataset. The normalized relative $L_2$ error is then defined as:
\begin{equation}
    \epsilon = \frac{||\hat{\phi}_{\text{pred}} - \hat{\phi}_{\text{true}}||_2}{||\hat{\phi}_{\text{true}}||_2},
\end{equation}
where $\hat{\phi}_{\text{pred}}$ and $\hat{\phi}_{\text{true}}$ denote standardized predicted and ground-truth flow fields, respectively. This normalization eliminates scale discrepancies and aligns with established practices in industrial \ac{cfd} model evaluation.

\begin{figure}[t!]
    \centering
    \includegraphics[width=\linewidth]{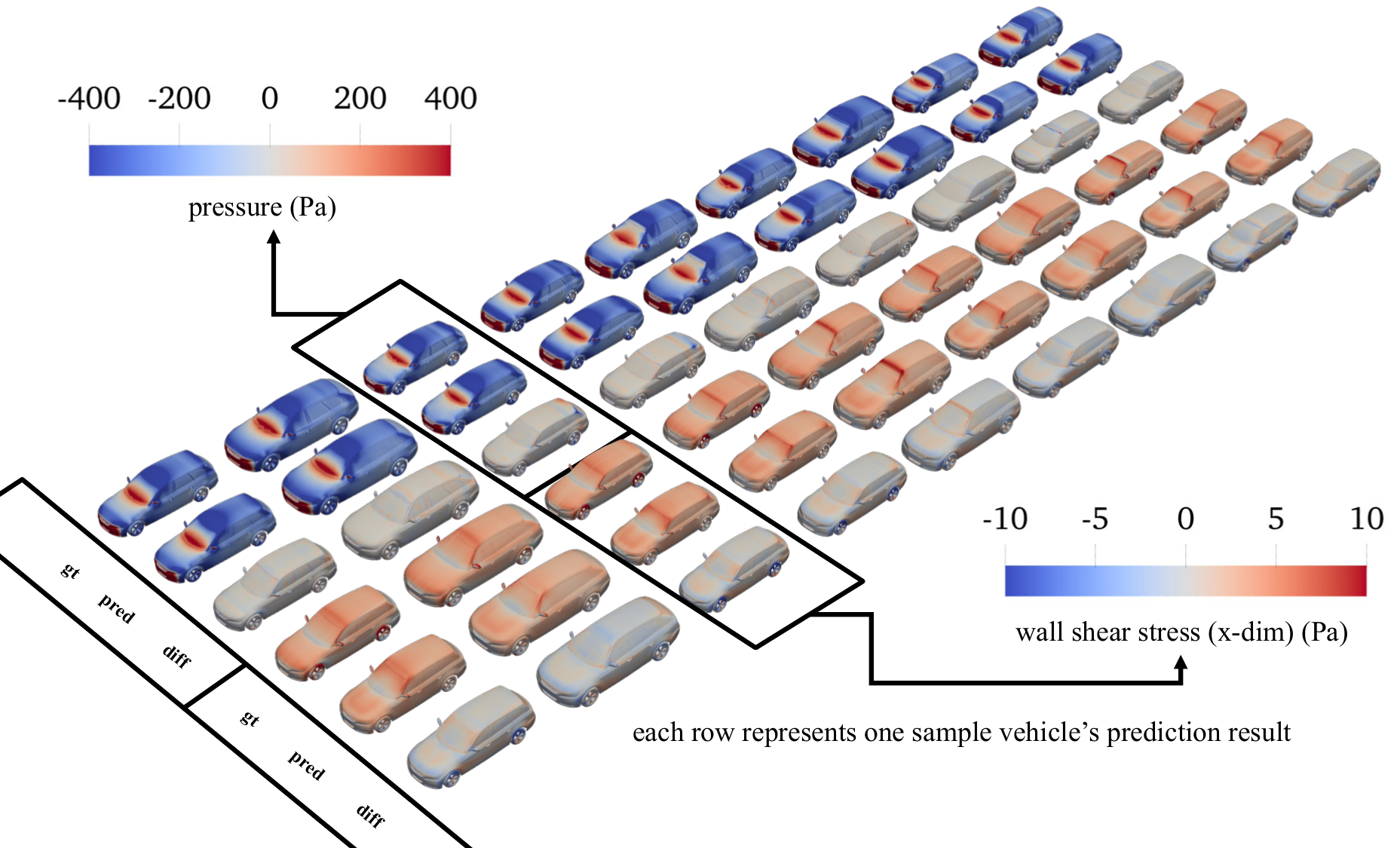} 
    \caption{\textbf{Benchmark results visualization.} Nine testing cases from the single-vehicle setup demonstrate model performance across unique geometric configurations. Each row presents pressure fields (left) and wall shear stress results (right), with ground truth \ac{cfd} data, machine learning predictions, and difference maps arranged from left to right within each prediction task.}
    \label{supp:fig:benchmark_park} 
\end{figure}

\paragraph{Drag Coefficient Counts Error}

Automotive industry practice quantifies drag coefficient accuracy using ``counts error,'' where 1 count equals 0.001 in $C_D$ units. For example, if the simulated $C_D$ is 0.300 and the predicted value is 0.320, the count error is 20 counts. This metric provides intuitive error quantification directly relevant to vehicle development targets.

\subsection{Benchmark Result}\label{supp:sec:benchmark_results}

\cref{tab:framework_comp} presents a systematic performance comparison between Transolver, GNOT, and PointNet frameworks across aerodynamic prediction tasks, revealing critical insights into their respective capabilities.

\paragraph{Single-Vehicle Specialization}

Transolver achieves superior performance in estateback-specific training, recording the lowest validation losses at 1,200 samples: pressure (0.1996), WSS (0.7953), and $C_D$ (0.0266). The framework demonstrates strong scalability, with $C_D$ prediction error decreasing 15.4\% when expanding training data from 400 to 1,200 samples. GNOT shows substantial improvement in high-data regimes, with 1,200-sample $C_D$ validation loss (0.0357) approaching Transolver's performance, indicating enhanced learning capacity with sufficient data. PointNet exhibits inconsistent convergence behavior, showing pressure validation loss degradation (3.1\% increase from 0.2991 to 0.2984) despite tripling training data. Single-vehicle benchmark results are visualized in \cref{supp:fig:benchmark_park}.

\paragraph{Multi-Vehicle Generalization}

Transolver maintains robust performance across vehicle configurations, showing minimal WSS validation loss degradation (2.9\%) between single-vehicle (0.7953) and 400×3 multi-vehicle (0.7871) scenarios. GNOT demonstrates superior cross-vehicle generalization, with 400×3 multi-vehicle $C_D$ validation loss (0.0315) outperforming single-vehicle baseline (0.0357) by 11.8\%, suggesting inherent architectural advantages for heterogeneous geometries. PointNet struggles with multi-task learning, exhibiting 41.7\% higher average WSS validation loss compared to Transolver in equivalent configurations.

\paragraph{Task-Specific Analysis}

Wall shear stress prediction presents the greatest challenge across all frameworks, with the highest relative errors observed (Transolver: 0.7953, GNOT: 0.8124, PointNet: 0.6560 at 400×3 validation). Drag coefficient predictions reveal framework-dependent characteristics: Transolver achieves ultra-low errors (0.0266) through specialized training, while GNOT's physics-informed architecture enables superior error consistency (±6.2\% variation) across vehicle configurations.

\paragraph{Sample Efficiency Analysis}

Transolver reaches 90\% peak performance with 800 samples, showing marginal improvement (<2\%) at 1,200 samples. GNOT requires 1,200 samples for comparable accuracy, while PointNet exhibits negative scaling beyond 800 samples, with 1,200-sample training yielding 4.3\% higher pressure loss than the 800-sample case, indicating architectural limitations in leveraging large datasets.

\begin{table}[t!]
    \centering
    \small
    \caption{\textbf{Training and validation loss comparison between Transolver and GNOT.} Loss curves for pressure, wall shear stress (WSS), and drag coefficient ($C_D$) prediction tasks. Training configuration: 500 samples per vehicle type (estateback, fastback, notchback) for 1,500 total samples. All loss values are normalized by initial conditions for comparative analysis.}
    \label{tab:framework_comp}
    \subcaptionbox{Transolver\label{supp:tab:transolver}}[\linewidth]{%
        \begin{tabular}{@{}llcccccc@{}}
            \toprule
            \multirow{2}{*}{Category} & \multirow{2}{*}{Configuration} & \multicolumn{3}{c}{Training Loss} & \multicolumn{3}{c}{Validation Loss} \\
            \cmidrule(lr){3-5} \cmidrule(lr){6-8}
              &  & Pressure & WSS & $C_D$ & Pressure & WSS & $C_D$ \\
            \midrule
            \multirow{3}{*}{Single-Vehicle} 
            & 400\texttimes Estateback &\round{0.211184114} & \round{0.784064233} & \round{0.030536702} & \round{0.217281744} & \round{0.788900793} & \round{0.031417966} \\
            & 800\texttimes Estateback &\round{0.205037236} & \round{0.779291868} & \round{0.02959924} & \round{0.208625659} & \round{0.783736885} & \round{0.028278392} \\
            & 1200\texttimes Estateback &\round{0.196936935} & \round{0.788128436} & \round{0.031452585} & \round{0.199577332} & \round{0.795262158} & \round{0.026573399} \\
            \midrule
            \multirow{3}{*}{Multi-Vehicle} 
            & 133\texttimes3 Vehicles & 0.2076 & 0.8050 & 0.0421 & 0.2194 & 0.8049 & 0.0375   \\
            & 266\texttimes3 Vehicles & 0.1967 & 0.8130 & 0.0335 & 0.2075 & 0.8170 & 0.0335   \\
            & 400\texttimes3 Vehicles & 0.1957 & 0.7973 & 0.0333 & 0.2074 & 0.7871 & 0.0286   \\
            \bottomrule
        \end{tabular}%
    }%
    \\%
    \subcaptionbox{GNOT\label{supp:tab:gnot}}[\linewidth]{%
        \begin{tabular}{@{}llcccccc@{}}%
            \toprule
            \multirow{2}{*}{Category} & \multirow{2}{*}{Configuration} & \multicolumn{3}{c}{Training Loss} & \multicolumn{3}{c}{Validation Loss} \\
            \cmidrule(lr){3-5} \cmidrule(lr){6-8}
              &  & Pressure & WSS & $C_D$ & Pressure & WSS & $C_D$ \\
            \midrule
            \multirow{3}{*}{Single-Vehicle} 
            & 400\texttimes Estateback & \round{0.3250207304954529} & \round{0.9618685245513916} & \round{0.11439066380262375} & \round{0.3250207304954529} & \round{0.9618685245513916} & \round{0.11439066380262375} \\
            & 800\texttimes Estateback & \round{0.218487665} & \round{0.79912656545639} &\round{0.0462791435420513} &\round{0.225448996} &\round{0.79858321} &\round{0.051631253}  \\
            & 1200\texttimes Estateback & 0.1960 & 0.8135 & 0.0032 & 0.2087 & 0.8217 & \round{0.03566}  \\
            \midrule
            \multirow{3}{*}{Multi-Vehicle} 
            & 133\texttimes3 Vehicles & 0.2069 & 0.7802 & 0.0301 & 0.2122 & 0.7795 & 0.0344    \\
            & 266\texttimes3 Vehicles & 0.1968 & 0.7909 & 0.0284 & 0.2002 & 0.7920 & 0.0309    \\
            & 400\texttimes3 Vehicles & 0.1954 & 0.8171 & 0.0282 & 0.1964 & 0.8124 & 0.0315    \\
            \bottomrule
        \end{tabular}%
    }%
    \\%
    \subcaptionbox{PointNet\label{supp:tab:pointnet}}[\linewidth]{%
        \begin{tabular}{@{}llcccccc@{}}%
            \toprule
            \multirow{2}{*}{Category} & \multirow{2}{*}{Configuration} & \multicolumn{3}{c}{Training Loss} & \multicolumn{3}{c}{Validation Loss} \\
            \cmidrule(lr){3-5} \cmidrule(lr){6-8}
              &  & Pressure & WSS & $C_D$ & Pressure & WSS & $C_D$ \\
            \midrule
            \multirow{3}{*}{Single-Vehicle}
            & 400\texttimes Estateback &   \round{0.280450} & \round{0.645958} & \round{0.062722} & \round{0.299060} & \round{0.654717504} & \round{0.050200883} \\  
            & 800\texttimes Estateback &   \round{0.254887} & \round{0.675544} & \round{0.037595} & \round{0.278658} & \round{0.676875889} & \round{0.049885195} \\  
            & 1200\texttimes Estateback &  \round{0.271736} & \round{0.615718} & \round{0.057094} & \round{0.298369} & \round{0.627303540} & \round{0.043648217} \\  
            \midrule
            \multirow{3}{*}{Multi-Vehicle} 
            & 133\texttimes3 Vehicles & 0.2971 &  0.6716 &  0.0722 &  0.3201 &  0.6767 &  0.0536  \\
            & 266\texttimes3 Vehicles & 0.2731 &  0.6875 &  0.0585 &  0.3007 &  0.7041 &  0.0653  \\
            & 400\texttimes3 Vehicles & 0.2872 &  0.6316 &  0.0589 &  0.3030 &  0.6560 &  0.0757  \\
            \bottomrule
        \end{tabular}%
    }%
\end{table}

\subsection{Scaling Analysis with Training Sample Size}\label{supp:sec:scaling}

We conducted comprehensive scaling experiments using Transolver in multi-vehicle configurations to assess training dataset size impact on drag coefficient prediction. The analysis utilized 12,000 total samples partitioned into training subsets of varying sizes, with 150 independently generated samples for testing. This revised data splitting strategy ensures robust evaluation, with minor discrepancies from initial results attributed to refined subset partitioning. Validation loss and relative improvement (calculated as $\text{Improvement} = \frac{\text{Loss}_{400} - \text{Loss}_{N}}{\text{Loss}_{400}} \times 100\%$ against the 400-sample baseline) are presented in \cref{table:cd_scaling}.

Results demonstrate consistent $C_D$ validation loss reduction with increased training data: doubling samples from 400 to 800 reduces loss by 10.67\%, scaling to 1,200 samples achieves 23.73\% reduction, and the full 12,000-sample dataset delivers 29.07\% total improvement. This trend confirms Transolver's effective data scaling properties, where larger training subsets directly translate to enhanced $C_D$ prediction accuracy. The substantial accuracy gains achieved with the 12,000-sample dataset provide meaningful improvements for practical \ac{cfd} applications, where marginal $C_D$ prediction error reductions drive significant advances in vehicle aerodynamic optimization workflows. Raw experimental data, including detailed loss curves and test sample distributions, are provided in supplementary materials to ensure reproducibility.

\section{Data Sources of Automotive \acs{cfd} Datasets}\label{supp:sec:vsbaseline}

The comparative metrics presented in \cref{tab:dataset_comparison} were compiled from multiple sources. For the DrivAerNet++ dataset~\citep{elrefaie2024drivaernetpp}, the wall $y^+$ range and the average $C_D$ precision were derived from the published simulation cases, while the wind tunnel error was extracted from Table~8 of the original paper. For the DrivAerML dataset~\citep{ashton2024drivaerml}, the wall $y^+$ range was not reported in the available literature; the average $C_D$ precision was estimated from Figure~13(a) of the original paper, and the wind tunnel error was obtained from Table~1 in its appendix.

\section{Experimental Setting and Details}\label{supp:sec:exp_detail}

\paragraph{Implementation Details}

To ensure experimental reproducibility, all random number generators were initialized with a fixed seed of 42. Models were trained and evaluated on the \dataset dataset under two configurations: 1 Vehicle (single configuration) and 3 Vehicle (multi-configuration). To assess data efficiency, the training set size was varied across 400, 800, and 1,200 samples, while the validation and test sets were fixed at 150 samples each. All models were trained from scratch for up to 500 epochs to ensure convergence. Training was conducted using the Adam optimizer with an initial learning rate of 0.001, adjusted by a cosine annealing scheduler, and a batch size of 4 and 1 for training and validation, respectively.

\paragraph{Network Architecture: Transolver}

The Transolver architecture~\citep{wu2024transolver} consists of 4 encoding layers, each implementing multi-head attention with 4 heads. The model uses a hidden dimension of 64 and maintains an MLP ratio of 1 in its feed-forward networks. The network processes 7-dimensional spatial input data and generates 4-dimensional outputs. The architecture incorporates 5 downsampling operations, utilizes a reference size of 8, and processes 16 data slices in parallel. During training, gradient clipping with a maximum norm of 0.1 was applied to stabilize optimization, while a weight decay of 0.0001 was used for regularization.

\paragraph{Network Architecture: PointNet}

The PointNet architecture~\citep{qi2017pointnet} processes irregular mesh data through a transformation network (T-Net) and hierarchical feature extraction. The input transformation network generates a $7 \times 7$ transformation matrix using three 1D convolutional layers with channel dimensions of 64, 128, and 256. The main network employs sequential 1D convolutional layers with batch normalization and ReLU activation, progressively increasing feature dimensions from 64 to 256 channels. A residual connection bridges the initial 64-channel features to the final representation, while point-wise fully connected layers map the 256-dimensional features to 4-dimensional outputs.

\paragraph{Network Architecture: GNOT}

The GNOT architecture~\citep{hao2023gnot} processes irregular mesh data using neural operator principles. The input transformation network processes 7-dimensional features through three 1D convolutional layers (64, 128, and 256 channels), followed by max pooling and fully connected layers to generate transformation matrices. The main network consists of sequential 1D convolutional layers with batch normalization and ReLU activation, progressively increasing feature dimensions from 64 to 128 and finally to 256 channels. A residual connection via 1×1 convolution bridges the initial 64-channel features to the final 256-channel representation. Point-wise fully connected layers conclude the architecture by mapping 256-dimensional features to the 4-dimensional output space.

\section{Limitations and Future Work}\label{supp:sec:limitations}

\paragraph{Geometric Representation}

While our dataset employs 20 deformation parameters, this parameterization remains insufficient to fully characterize complete vehicle geometries. The simplified geometric representation limits the model's ability to capture fine-scale aerodynamic features that significantly impact flow behavior in real-world applications.

\paragraph{Simulation Fidelity}

Although our simulations incorporate many industrial practices, they cannot capture all real-world complexities, including dynamic effects, environmental conditions, and manufacturing tolerances. The steady-state \ac{rans} approach, while computationally efficient, may not fully resolve transient flow phenomena critical for certain aerodynamic assessments.

\paragraph{Model Performance Validation}

Our experiments confirm that data-driven scaling laws remain effective, with model performance improving consistently as training samples increase. However, the observed improvements may plateau beyond current dataset sizes, requiring investigation of alternative approaches for continued accuracy gains.

\paragraph{Future Directions}

Future work will focus on dataset enhancement through expanded geometric parameterization and foundation model development for industrial applications. We plan to explore physics-informed learning methods and data assimilation techniques incorporating wind tunnel experimental data to improve model performance in limited-data scenarios. Additionally, investigation of multi-fidelity approaches combining \ac{rans}, LES, and experimental data may enhance prediction accuracy while maintaining computational efficiency for practical automotive design workflows.

\clearpage
\begin{figure}[t!]
    \centering
    \includegraphics[width=0.95\linewidth]{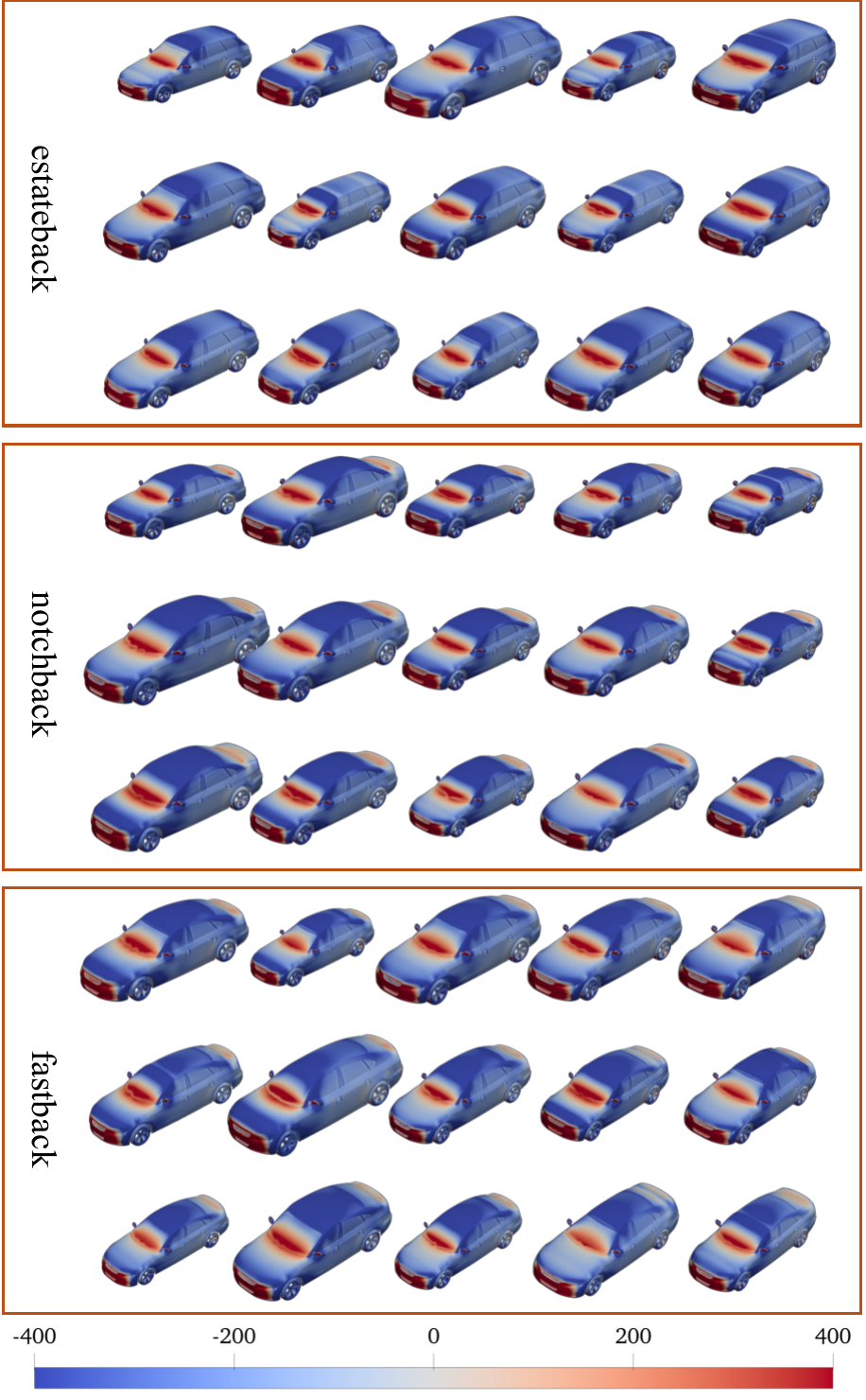} 
    \caption{\textbf{\dataset simulation results: pressure field distributions (Pa).}}
    \label{supp:fig:pressure_park} 
\end{figure}
\clearpage
\begin{figure}[t!]
    \centering
    \includegraphics[width=0.95\linewidth]{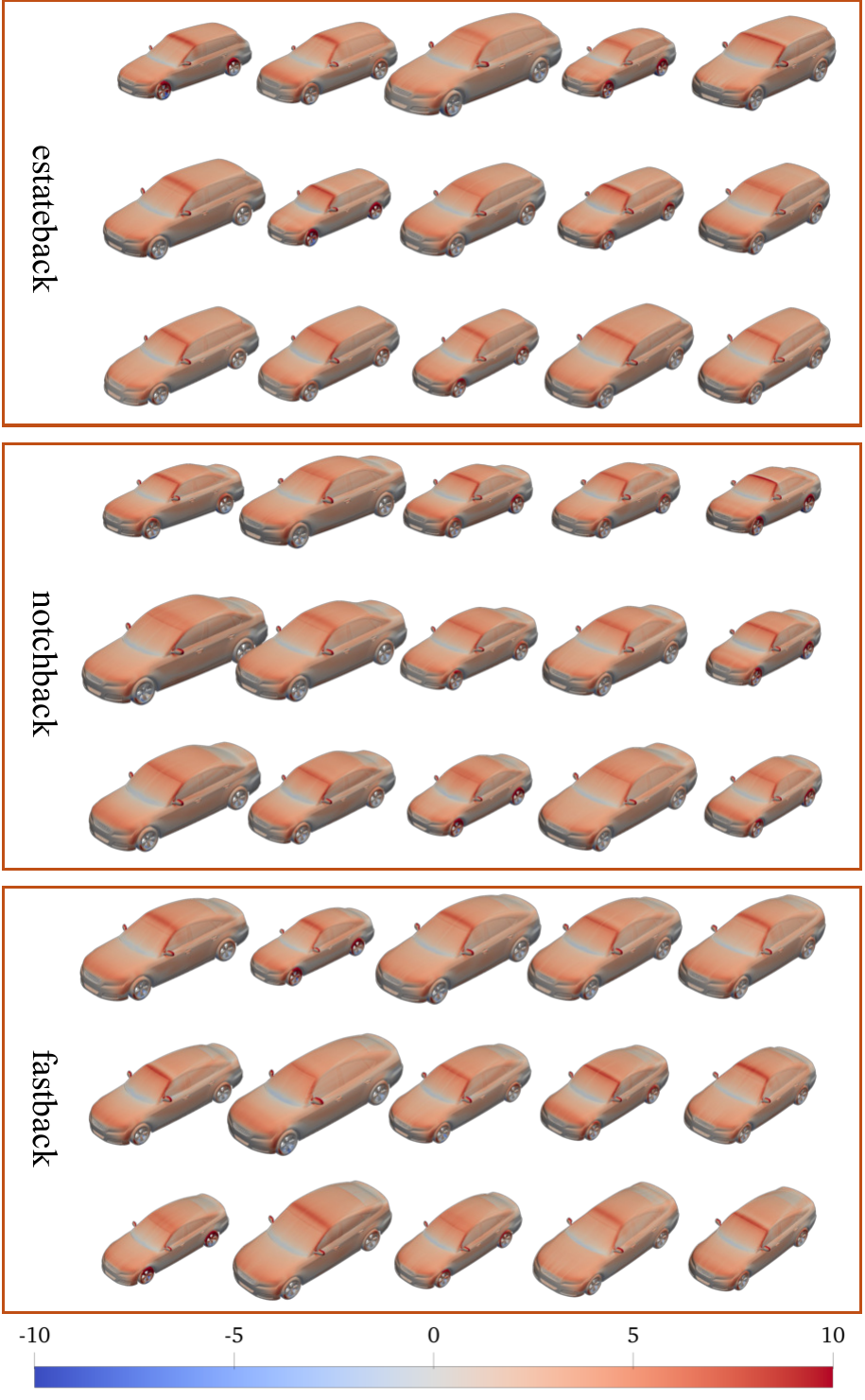} 
    \caption{\textbf{\dataset simulation results: wall shear stress distributions (Pa).}} 
    \label{supp:fig:wss_park} 
\end{figure}
\clearpage
\begin{figure}[t!]
    \centering
    \includegraphics[width=0.95\linewidth]{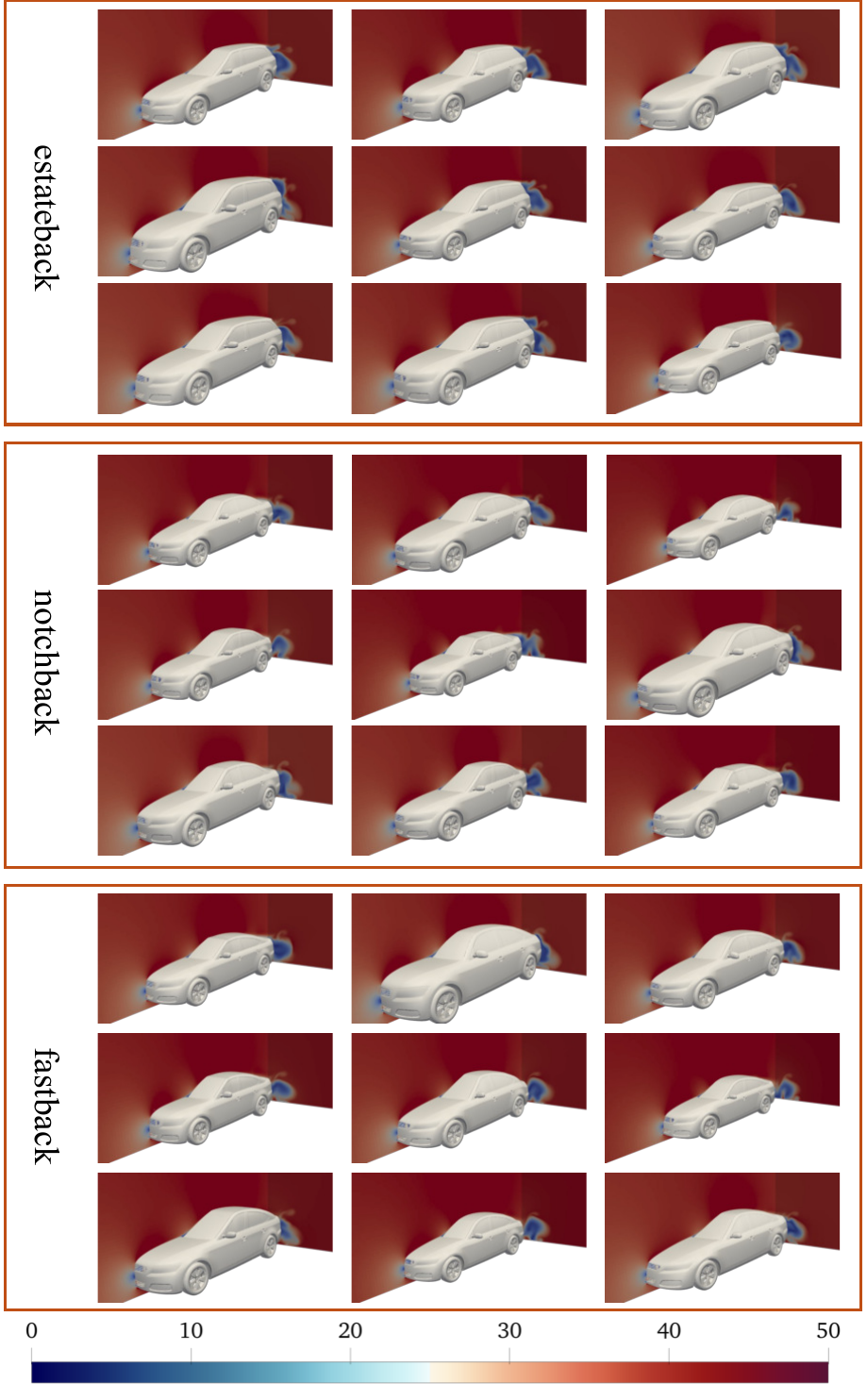} 
    \caption{\textbf{\dataset simulation results: velocity field distributions (m/s)}} 
    \label{supp:fig:velocity_park} 
\end{figure}

\end{document}